\pdfoutput=1

\documentclass[remotesensing,article,accept,moreauthors]{Definitions/mdpi}

\usepackage{amsfonts}
\usepackage{nicefrac}
\usepackage{threeparttable}
\usetikzlibrary{positioning,arrows.meta,fit,backgrounds}

\firstpage{1} 
\pubvolume{18}
\issuenum{22}
\articlenumber{3975}
\pubyear{2026}
\copyrightyear{2026}
\externaleditor{Jianhong Liu, Clement Atzberger, Peijun Sun, Xianfeng Liu and Donghai Zhang}
\datereceived{5 October 2026} 
\daterevised{12 November 2026} %
\dateaccepted{16 November 2026} 
\datepublished{19 November 2026} 

\Title{Benchmarking Geospatial Foundation Models Against a Supervised Baseline for Fine-Grained Alpine Habitat Change Detection with Very High-Resolution Imagery and LiDAR $^{\dagger}$}

\Author{Harald Kristen %
 $^{1,2}$\orcidA{}, Daniel Kulmer $^{1,2}$\orcidB{} and Manuela Hirschmugl $^{1,2,}$*\orcidC{}}

\AuthorNames{Harald Kristen, Daniel Kulmer and Manuela Hirschmugl}

\address{%
$^{1}$ \quad Department of Geography and Regional Science, University of Graz, %
  8010 Graz, Austria; harald.kristen@joanneum.at (H.K.); daniel.kulmer@joanneum.at (D.K.) %
\\
$^{2}$ \quad Institute DIGITAL, JOANNEUM RESEARCH, 8010 Graz, Austria}%

\corres{Correspondence: manuela.hirschmugl@uni-graz.at}

\conference{Kulmer, D.; Kristen, H.; Hirschmugl, M. Habitat/Land Cover Change Detection in Alpine Protected Areas: A Comparison of AI Architectures. In Proceedings of the AGIT 2026 Conference for Geoinformatics, Salzburg, Austria, 8--9 July 2026}%

\abstract{Geospatial foundation models (GFMs) are proposed for operational Earth observation but benchmarked mostly on urban and agricultural data, so their value and limitations for fine-grained ecological monitoring, with fuzzy class boundaries and severe class imbalance, are largely untested. We benchmark five GFMs (Clay v1.0, Clay v1.5, Prithvi-EO-2.0, DOFA, TerraMind) against two supervised U-Net baselines, one with a transformer (MiT-B2) and one with a convolutional (ResNet-50) encoder, for post-classification habitat change detection in Ges\"ause National Park, Austria. All models are trained with three seeds on 2013 expert-interpreted labels (23 habitat classes), using very high-resolution red, green, blue (RGB) and near-infrared aerial imagery and a normalised digital surface model (nDSM) derived from airborne light detection and ranging (LiDAR). Applied to 2019 imagery and 2020 LiDAR, their predictions are compared with the 2013 reference to yield change maps with eight transition classes plus no change. All five GFMs are robustly ahead of the convolutional U-Net, but so is the transformer U-Net, which only Clay v1.0 exceeds, by a small margin (multi-class change macro Intersection over Union (IoU) 0.233 vs. 0.209) whose paired 95\% confidence interval (+0.001 to +0.045) nearly includes zero. Adding the nDSM helps every model, but mostly in the canopy-cover classes, whose reference labels are themselves LiDAR-derived. Rare or heterogeneous transitions remain poorly detected by every model. Rule-based post-processing defined with the Ges\"ause National Park administration converts the predictions into polygon-structured, ecologically plausible change maps that meet the park's data standards, with little effect on pixel accuracy (multi-class macro IoU $-0.03$ to +0.04) but lower recall for forest density loss and setback; the maps are therefore suited to flagging candidate changes for expert verification rather than to unsupervised use.}

\keyword{geospatial foundation models; change detection; habitat monitoring; LiDAR; class imbalance; alpine protected areas}

\addhighlights{yes}
\renewcommand{\addhighlights}{\\%
\textbf{What are the main findings?}
 \begin{itemize}[labelsep=2.5mm,topsep=-3pt]
 \item All five geospatial foundation models clearly beat a convolutional U-Net in Alpine habitat change detection, but a U-Net with a transformer encoder matches four of them.
 \item LiDAR height data improve change detection for every model, mostly by separating canopy-cover levels.
 \end{itemize}\vspace{3pt}
\textbf{What are the implications of the main findings?}
 \begin{itemize}[labelsep=2.5mm,topsep=-3pt]
 \item With full training data, a supervised U-Net remains a strong reference for habitat monitoring; Clay v1.0, partly pretrained on aerial imagery, outperforms it only slightly, at about 45 times the computation per patch at inference.
 \item The open HabitAlp~2.0 dataset on Hugging Face adds fine-grained ecological habitats to foundation-model benchmarks, which have so far focused on urban and agricultural~tasks.
 \end{itemize}
}

\begin{document}

\section{Introduction}
\label{sec:intro}

Alpine protected areas are important biodiversity reservoirs and early indicators of climate change \cite{Hauenstein2013CC}. They experience rapid landscape transformation through climatic shifts and geomorphological processes such as rockfall, floods, landslides, and avalanches \cite{Pauli2012Recent}, as well as insect outbreaks, wildfires, and storms. With Alpine ecosystems warming at roughly twice the global average rate \cite{European2009Regional}, frequent and accurate habitat monitoring is essential for conservation management.

For this study we adopt the habitat definition of Bunce et al. \cite{Bunce2008standardized}: an ``element of the land surface that can be consistently defined spatially in the field in order to define the principal environments in which organisms live''%
. Traditional Alpine habitat mapping relies on manual stereo interpretation of aerial imagery, exemplified by the INTERREG ``HabitAlp'' project~\cite{Hauenstein2013CC}. While this produces detailed baseline data, it is labour-intensive and costly, and manual updates cannot keep pace with the monitoring frequency that proactive management~requires.

High-resolution remote sensing combined with deep learning potentially offers a path toward automated mapping \cite{Lu2024AI}. In particular, geospatial foundation models (GFMs) such as Prithvi-EO-2.0, Clay, DOFA, and TerraMind are pretrained with self-supervision on large Earth-observation archives and fine-tuned for downstream tasks, which in principle should reduce dependence on labelled data \cite{Szwarcman2025Prithvi, Clay2023Clay, Xiong2024DOFA, Jakubik2025TerraMind, Wang2022Self}. However, GFMs have so far been evaluated overwhelmingly on clearly separable land-cover classes (agriculture, roads, built-up) and on urban or agricultural benchmark datasets \cite{Lacoste2023GEOBench, Marsocci2024PANGAEA}. How they behave on fine ecological habitat types with fuzzy boundaries, gradual transitions, and severe class imbalance has, to our knowledge, not been tested. Recent work also cautions that complex architectures do not always beat well-tuned baselines \cite{Corley2024Change}. %
Reported scores for the same model and dataset diverge between papers because evaluation protocols differ \cite{Corley2026SOTA}, so we retrain and evaluate every model here under one protocol.

Using part of the HabitAlp dataset, a long-term, expert-interpreted habitat mapping of Ges\"ause National Park, Austria \cite{Hoffert2006Digitale, Hauenstein2018CC}, we benchmark GFMs against two strong supervised U-Net baselines for habitat change detection. We focus on the \emph{post-classification} paradigm, in which habitat maps are compared between dates \cite{Colditz2012Potential}. Here, a model trained on the existing reference map predicts a habitat map for the later date, which is then compared with that reference. Our contributions answer four research questions:

\begin{enumerate}[label=\emph{RQ\arabic*}, leftmargin=*, labelsep=8pt]
    \item How well can current deep-learning methods detect fine-grained habitat change in an Alpine protected area, and which transitions are reliably detectable? %
    \item Do geospatial foundation models outperform strong supervised baselines? %
    \item What is the added value of LiDAR-derived height information for single-date segmentation and for change detection?
    \item Can rule-based post-processing turn pixel-level change predictions into a product usable by park management, and at what cost in pixel-level accuracy?
\end{enumerate}

\textls[-15]{The main contributions of this article are the following: (1) an open, expert-labelled benchmark for fine-grained habitat change detection with 23 habitat classes, 20\,cm RGB and near-infrared orthophotos, LiDAR nDSM, and an independent cross-temporal test set \cite{Kristen2025HabitAlpData}; (2) an evaluation of five GFMs against two supervised U-Net baselines on fine-grained ecological classes, a setting absent from existing GFM benchmarks, with all models trained with three seeds under one protocol and compared with paired confidence intervals over test tiles and seeds; (3)~an input ablation that quantifies the value of LiDAR height for each architecture, for both single-date segmentation and change detection; and (4) a rule-based post-processing pipeline that converts pixel predictions into a physically plausible, polygon-structured product usable by park administrations. Code and data are openly available \cite{Kristen2025HabitAlpCode, Kristen2025HabitAlpData}.}

This article is a revised and expanded version of a paper entitled ``Habitat/Land Cover Change Detection in Alpine Protected Areas: A Comparison of AI Architectures'', which was presented at the AGIT 2026 Conference for Geoinformatics, Salzburg, Austria, 8--9 July 2026 \cite{Kulmer2026AGIT}. Relative to that work, we drop the direct change-detection paradigm (no multimodal inputs at the historical date) and expand the post-classification benchmark with: full documentation of the class scheme, transition rules, training protocol, and per-model band adaptation (Section~\ref{sec:methodology}, Appendix~\ref{app:classes}); class-balanced metrics (macro IoU/F1, balanced accuracy) replacing overall accuracy, with confidence intervals (CIs) from a paired bootstrap over test tiles and training seeds; a convolutional ResNet-50 U-Net baseline and a decoder check; per-class results for all models (Appendix~\ref{app:perclass}); a single-date segmentation and LiDAR ablation; an operational post-processing pipeline; and an extended discussion of pretraining data composition. All models were retrained for this article with three seeds each, on a split that masks the cross-temporal test area, and with a fixed budget of 200~epochs, so all numbers differ from those of the conference version; the MiT-B2 U-Net loses most and Prithvi-EO-2.0 gains most.

\subsection{Related Work}
\label{sec:related}

\textls[-15]{Methodologically, change detection has moved from simple spectral differencing toward machine-learning and object-based approaches \cite{Miranda2022Monitoring}, and within deep learning from CNNs to transformers and, most recently, large pretrained foundation models \cite{Bai2023Deep, Janga2023Review}. Independent of this trend, two paradigms exist: \emph{post-classification} change detection produces an independent map per date and compares them, at the cost of potential error propagation \cite{Colditz2012Potential}; \emph{direct} change detection feeds bi-temporal imagery to a single model trained to predict change end-to-\linebreak end \cite{Parelius2023Review, Zhu2024ChangeViT}. Both are now typically implemented with deep-learning backbones. In our post-classification setting, only the later map is predicted and compared with the existing reference, so errors enter from one date only. We adopt post-classification because it reuses pretrained segmentation backbones and accommodates differing modalities across acquisition dates: LiDAR and colour-infrared are not available for every historical year (Table~\ref{tab:data_availability}), so each date can be mapped from whatever sensors it has, whereas a direct bi-temporal model would require an identical multimodal input stack at both dates.}

\textbf{Supervised baselines. %
} U-Net remains a dominant and hard-to-beat segmentation architecture \cite{Ronneberger2015U}, and well-configured U-Nets frequently match or exceed more complex models \cite{Corley2024Change}. We therefore use U-Nets with ImageNet-pretrained encoders (MiT-B2 and ResNet-50) as the reference against which GFMs must show improvement. %

\textls[-15]{\textbf{Geospatial foundation models.} GFMs learn transferable representations from large unlabelled archives via self-supervised pretext tasks such as masked reconstruction \cite{Wang2022Self}. They differ markedly in pretraining data: Prithvi-EO-2.0 is trained on 4.2\,million Harmonized Landsat Sentinel-2 (HLS; 30\,m) samples \cite{Szwarcman2025Prithvi}; Clay additionally ingests sub-metre aerial imagery (NAIP, LINZ) alongside satellite data \cite{Clay2023Clay, U2025NAIP, Land2025LINZ}; DOFA introduces wavelength-conditioned encoders pretrained across five sensors including NAIP aerial imagery ($\sim$1\,m) \cite{Xiong2024DOFA}; and TerraMind is a generative multimodal model \cite{Jakubik2025TerraMind}. %
These differences, particularly the resolution gap between satellite-only and aerial pretraining, motivate testing whether pretraining data composition matters for very high-resolution ecological tasks. Scale-MAE \cite{Reed2023ScaleMAE} shows that making pretraining explicitly resolution-aware improves transfer across ground sampling distances, the problem satellite-pretrained encoders face on sub-metre imagery. AlphaEarth Foundations \cite{Brown2025AlphaEarth} and TESSERA \cite{Feng2025TESSERA} are not fine-tunable encoders but release precomputed annual embeddings (at 10\,m), and neither covers our 2013 training epoch (AlphaEarth starts in 2017, and TESSERA relies on Sentinel-2, which was launched in 2015), so they could not be included in the benchmark. Benchmark studies increasingly question whether GFMs deliver on downstream tasks: GEO-Bench \cite{Lacoste2023GEOBench} and PANGAEA \cite{Marsocci2024PANGAEA} find that no single GFM dominates and that supervised baselines remain competitive on several tasks, and Nedungadi et al. \cite{Nedungadi2026Agriculture} report unstable model rankings and a gap between pretraining and deployment modalities when GFMs are applied to agricultural problems. Zhou et al. \cite{Zhou2026Spatial} review the spatial explicitness of GeoAI and geospatial foundation models and outline a roadmap for its evaluation. Spatial explicitness is the degree to which a model represents and uses location, spatial context, and spatial dependence instead of treating samples as independent. Fine-grained ecological habitat classes are absent from all of these benchmarks.}

\textbf{Ecological monitoring.} Deep-learning change detection in ecology has largely addressed simple, few-class problems such as forest-cover loss \cite{Miranda2022Monitoring, Sadel2025Monitoring, Tharun2024Deforestation}, while studies achieving high accuracy on coarse vegetation classes do not readily extend to fine, imbalanced habitat schemes \cite{kattenborn_convolutional_2019, shi_evaluation_2025}. Adapting GFMs to such settings requires domain-specific fine-tuning and effective multimodal fusion while keeping the workflow operationally feasible \cite{Li2024New, Ding2024Adapting, Tuia2022Seeing}. Ghamisi et al. \cite{Ghamisi2025Geospatial} argue for evaluation beyond raw accuracy, which motivates our operational focus.
\begin{table}[H]
\caption{Data in the HabitAlp~2.0 dataset for the study area and their use in this study: T = training epoch (2013 imagery and labels with 2010/11 LiDAR; training, validation, and in-domain test blocks), E = cross-temporal test epoch (2019 imagery and labels with 2020 LiDAR, ``2019/20'' in the text), \checkmark\ = available but not used.\label{tab:data_availability}}%
\tablesize{\small}
\begin{tabularx}{\textwidth}{llCCCCC}
\toprule
\textbf{Data Type} & \textbf{Resolution} & \textbf{2003/04 }& \textbf{2010/11} & \textbf{2013} & \textbf{2019} & \textbf{2020} \\
\midrule
RGB orthophoto & 20 cm & \checkmark & - & T & E & - \\
CIR orthophoto (NIR band) & 20 cm & - & - & T & E & - \\
LiDAR nDSM & 1 m & - & T & - & - & E \textsuperscript{a}  \\
Habitat reference labels & polygons & \checkmark & - & T & E \textsuperscript{b}  & - \\
\bottomrule
\end{tabularx}
\noindent{\footnotesize{\textsuperscript{a} Four patches only (11.9\,km$^2$), which define the cross-temporal test area. \textsuperscript{b} Interpreted on the 2019 orthophotos; referred to as the 2019/20 reference in the text because the paired nDSM dates from 2020.}}

\end{table}
\section{Materials and Methods}
\label{sec:methodology}

\subsection{Study Area}

\begin{figure}[H]
    \includegraphics[width=\textwidth]{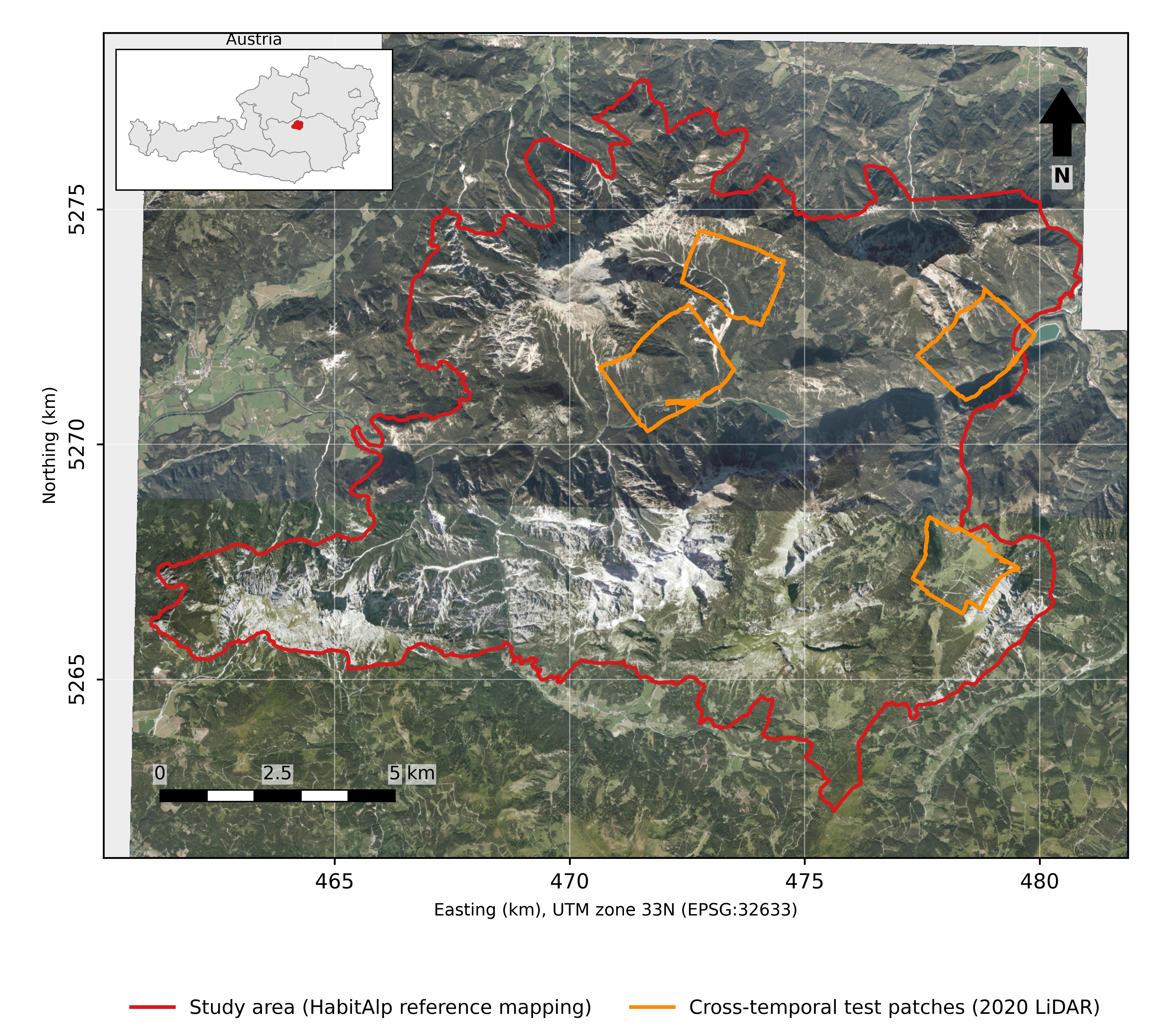}
    \caption{Study area in Styria, Austria: %
 the extent of the HabitAlp reference mapping in and around Ges\"ause National Park (red, $\sim$154\,km$^2$), which provides the training, validation, and in-domain test blocks. The four orange patches are the cross-temporal test patches, defined by the 2020 LiDAR coverage. Background: 2019 orthophoto \cite{2023Digitale}; coordinates in UTM zone 33N.\label{fig:study_area}}
\end{figure}

The study area (Figure~\ref{fig:study_area}) is the $\sim$154\,km$^2$ extent of the HabitAlp reference mapping in and around Ges\"ause National Park in Styria, Austria, a site of the European Long-Term Ecosystem Research (eLTER) network \cite{Lippl2024Benchmark}. The landscape combines extensive forests, dynamic waterways, alpine meadows, and rocky terrain \cite{Kirchmeir2023Methodenentwicklung}. Roughly 70\% of the park's forest is protected biotope forest where natural processes such as bark-beetle infestation proceed without intervention \cite{NationalparkBiotopschutzwald}, producing ecological dynamics rarely found elsewhere in Austria.

\subsection{Remote Sensing and Reference Data}

All data are published as the open HabitAlp~2.0 dataset \cite{Kristen2025HabitAlpData}, which bundles, for the study area, 20\,cm aerial orthophotos (RGB for 2003/04, 2013, and 2019; colour-infrared (CIR) for 2013 and 2019) from GIS Steiermark \cite{2023Digitale}, airborne LiDAR products at 1\,m \cite{2022Digitales}, the HabitAlp habitat reference for 1954, 2003, and 2013, and reference change maps. This study uses the subset needed for a multimodal change benchmark (Table~\ref{tab:data_availability}): RGB, the near-infrared (NIR) band of the CIR orthophotos, and the LiDAR-derived normalised digital surface model (nDSM, height above ground), at two epochs. The \emph{training epoch} pairs the 2013 orthophotos and reference labels with the 2010/11 LiDAR, which covers the full extent. The resulting 1--3 year mismatch between nDSM and imagery is small relative to the slow change of canopy structure in this Alpine setting. We refer to this epoch as ``2013''. The \emph{test epoch} pairs the 2019 orthophotos, and reference labels interpreted on them, with 2020 LiDAR, also from GIS Steiermark. %
We refer to this epoch and its reference as ``2019/20'' throughout and use 2019 or 2020 only where the imagery or the LiDAR alone is meant. The 2020 LiDAR covers only four small areas, so the cross-temporal test set is limited to these four patches (approximately 11.9\,km$^2$). The 2003/04 imagery and the earlier reference years are part of the dataset but not used here. For training and inference, all layers are resampled on the fly to a common 0.5\,m grid (orthophotos downsampled, nDSM upsampled). Predictions are generated at 0.5\,m and resampled to the 0.2\,m grid of the rasterised reference maps by nearest-neighbour assignment before any metric is computed, so all segmentation and change metrics are pixel-based at 0.2\,m; pixels that are no-data in the reference or the prediction are excluded. All data use UTM zone 33N (EPSG:32633).

Reference labels come from the HabitAlp dataset (154\,km$^2$; 30{,}241 polygons) produced by expert stereo interpretation for 1954, 2003, and 2013 \cite{Hoffert2006Digitale, Hauenstein2018CC}. For 2019/20, the 2013 polygon delineation was carried forward and each polygon was re-interpreted on the 2019 orthophotos to determine whether a habitat transition had occurred, splitting polygons into sub-polygons with new classes where a change affected only part of a polygon. This interpretation was performed at the University of Graz and cross-checked by Nationalpark Ges\"ause domain experts. The original 115-class scheme was consolidated into 23 habitat classes balancing class frequency against forest-structural detail (Table~\ref{tab:classes}). For change detection, class-to-class transitions are aggregated into eight ecologically meaningful categories plus ``no change'' (Table~\ref{tab:transitions}), reducing label sparsity for very rare transitions. The ``other transition'' category is a residual class that collects every class-to-class change not assigned to one of the eight named categories, mostly because these transitions are rare. Among them are tree-species composition changes (coniferous\,$\leftrightarrow$\,broad-leaved), conversions between forest and non-forest cover (grassland, bare ground, water), and transitions whose target class is undefined in the reference.

\begin{table}[H]
\caption{Consolidated habitat classes and their area share in the 2013 reference over the full study area (CC = canopy cover, \%). Shares refer to the labelled area; 141.6\,ha of clearcut masked as no-data (Appendix~\ref{app:classes}) are excluded.\label{tab:classes}}

\begin{tabularx}{\textwidth}{lRR}
\toprule
\textbf{Class Name} & \textbf{Area (ha)} & \textbf{Share (\%)} \\
\midrule
Rock & 2632.8 & 17.26 \\
Mountain dwarf forest (`Krummholz' belt) & 2225.5 & 14.59 \\
Coniferous mature forest CC $\geq$ 80 & 1667.1 & 10.93 \\
Alpine grassland, heath & 1095.7 & 7.18 \\
Coniferous mature forest CC < 80 & 1010.7 & 6.63 \\
Old coniferous forest, multilayered CC < 80 & 969.0 & 6.35 \\
Coniferous pole timber CC $\geq$ 80 & 787.6 & 5.16 \\
Old coniferous forest, multilayered CC $\geq$ 80 & 695.8 & 4.56 \\
Debris-covered areas & 594.7 & 3.90 \\
Broad-leaved mature forest CC $\geq$ 80 & 539.3 & 3.54 \\
Grassland, buffer strip forest/open land & 462.7 & 3.03 \\
Erosion area, gully & 447.3 & 2.93 \\
Old broad-leaved forest, multilayered CC $\geq$ 80 & 439.8 & 2.88 \\
Young coniferous (growth, thicket) & 439.8 & 2.88 \\
Coniferous pole timber CC < 80 & 239.0 & 1.57 \\
Clearcut areas & 224.1 & 1.47 \\
Young broad-leaved (growth, thicket) & 206.1 & 1.35 \\
Broad-leaved pole wood & 161.3 & 1.06 \\
Other minor classes & 130.3 & 0.85 \\
Broad-leaved mature forest CC < 80 & 97.0 & 0.64 \\
Waterbody & 95.3 & 0.62 \\
Old broad-leaved forest, multilayered CC < 80 & 77.9 & 0.51 \\
Gravel bank, shoal, fluviatile & 17.2 & 0.11 \\
\bottomrule
\end{tabularx}
\end{table}
\vspace{-6pt}
\begin{table}[H]
\caption{Change transition classes and their area share in the 2013 to 2019/20 reference change map over the evaluated cross-temporal test area.\label{tab:transitions}}
\begin{tabularx}{\textwidth}{lRR}
\toprule
\textbf{Transition Class} & \textbf{Area (ha)} & \textbf{Share (\%)} \\
\midrule
No change & 683.8 & 58.9 \\
Forest stage progression & 264.4 & 22.8 \\
Mature tree density loss & 86.7 & 7.5 \\
Other transition & 36.5 & 3.1 \\
Old-growth density loss & 33.0 & 2.8 \\
Forest setback/young loss & 22.6 & 1.9 \\
New clearcut & 17.9 & 1.5 \\
Early forest establishment & 8.9 & 0.8 \\
Forest density gain & 6.4 & 0.6 \\
\bottomrule
\end{tabularx}
\end{table}

The study area is partitioned into a $12 \times 12$ regular grid ($\sim$1.5\,km cells) whose cells are randomly assigned to training (70\%), validation (15\%), and test (15\%) with a fixed seed, so validation and test data occupy spatially separate $\sim$1.5\,km blocks, disjoint from the training blocks.
For the main benchmark, the four 2019/20 test patches plus a 128\,m buffer (one training patch) are additionally masked out of the training and validation areas (masked split), so no training or validation patch overlaps the cross-temporal test area; this removes about 9\% of the labelled training and validation area. Because spatial autocorrelation between training and evaluation samples inflates accuracy estimates in remote sensing \cite{Karasiak2022Spatial, Kattenborn2022Spatially, Corley2026SOTA}, the mask keeps every training and validation patch at least 128\,m from the cross-temporal test area. The grid assignment and thus the 2013 test blocks are unchanged. As controls, the MiT-B2 U-Net is also trained with three seeds on the original split and on a split that drops every grid block touching a 2019/20 test patch (block-drop~split). %

\subsection{Models and Experimental Design}\label{sec:models} %

We compare five GFMs against two supervised U-Net baselines, all used as semantic-segmentation backbones in the same post-classification pipeline (Figure~\ref{fig:workflow}; Table~\ref{tab:hyperparameters}). The main baseline is a U-Net \cite{Ronneberger2015U} with a Mix Vision Transformer encoder (\texttt{mit\_b2} \cite{Xie2021SegFormer}, ImageNet-pretrained) from \texttt{segmentation\_models\_pytorch}, trained through the same TerraTorch pipeline as the GFMs. TorchGeo \cite{Stewart2024TorchGeo} provides the geospatial datasets and samplers for all models. Although the encoder is transformer-based rather than purely convolutional, we refer to this model as ``MiT-B2 U-Net'' and treat it as the main supervised baseline because it follows standard supervised ImageNet pretraining rather than the self-supervised Earth-observation pretraining that defines GFMs. As a purely convolutional second baseline, we add a U-Net with an ImageNet-pretrained ResNet-50 encoder (``ResNet-50 U-Net''), one of the segmentation baselines of GEO-Bench \cite{Lacoste2023GEOBench}, whose encoder is about as large as MiT-B2 and which is trained in the same way. The GFMs (Clay v1.0, Clay v1.5, Prithvi-EO-2.0 (300M), DOFA, and TerraMind) are fine-tuned through TerraTorch \cite{Gomes2025TerraTorch} under a common training protocol (AdamW, weight decay 0.05, ten-epoch linear warm-up followed by cosine annealing; see Training protocol below). For most models, the learning rate was first estimated with a learning-rate range test \cite{Smith2017Cyclical} and then checked by validation macro IoU in preliminary runs (Table~\ref{tab:r1-b8_lr_table}); the decoder was fixed per model (FCN for Clay and Prithvi-EO-2.0, U-Net for DOFA and TerraMind), so no systematic grid search was run. Appendix~\ref{app:model-selection} describes how the five GFMs were selected. %
To check whether the decoder drives the ranking, Clay v1.0 and Prithvi-EO-2.0 are also trained with a U-Net decoder (one seed each). All encoders are fully fine-tuned (not frozen).

\textbf{Training protocol.} All models are trained with cross-entropy loss weighted by inverse class area (weights normalised to sum to the number of classes, computed once from the full 2013 reference raster and used unchanged for the masked split) to counteract the severe class imbalance. Data augmentation includes random horizontal and vertical flips, random resized crop (scale 0.8--1.0), colour jitter (brightness 0.3, contrast 0.5), and random Gaussian blur. Inputs are normalised by per-band mean and standard deviation computed from the full training rasters. Each epoch consists of 512 randomly sampled training batches with a batch size of 4 for all models (2{,}048 patches per epoch). Model selection uses the macro IoU on 64 randomly sampled validation batches (256 patches) per epoch. Every model is trained for a fixed 200 epochs without early stopping (cosine annealing over the 190 epochs after warm-up), and the checkpoint with the highest validation macro IoU is used. Because the validation patches are a new random draw each epoch, best-epoch selection is noisy, but on the test patches the last epoch differs from the best checkpoint by only $-0.007$ to +0.009 macro IoU in segmentation (Table~\ref{tab:r1-b9_best_vs_last}; training facts per run in Table~\ref{tab:r1-b8_training_facts}). Each model is trained with three training seeds (0, 1, 2) on the masked split; the seeds fix initialisation and patch sampling, while GPU kernels are not bit-deterministic. Training uses 16-bit mixed precision. All experiments ran in Python~3.14 with PyTorch~2.10, TorchGeo~0.9, and Albumentations~1.4; these runs used TerraTorch~1.2.13. The full environment specification is in the code repository. The LiDAR ablation and post-processing analyses use the original runs on the unmasked split, trained once per model without a fixed training seed, with a batch size of 4 for the GFMs and 8 for the MiT-B2 U-Net, and up to 150 epochs with early stopping (patience 40 epochs, minimum 10 epochs); they used a TerraTorch 1.2.4 pre-release, which gives numerically identical model outputs to version 1.2.13.

\begin{figure}[H]
\begin{tikzpicture}[x=1cm, y=1cm, font=\scriptsize,
  box/.style={draw=black!60, rounded corners=1.5pt, align=center, inner sep=3pt, minimum height=7.5mm},
  data/.style={box, fill=blue!10},
  proc/.style={box, fill=orange!15},
  res/.style={box, fill=green!15},
  grp/.style={draw=black!40, dashed, rounded corners=3pt, inner sep=5pt, fill=gray!6},
  glab/.style={font=\scriptsize\bfseries, anchor=north west, inner sep=2pt},
  arr/.style={-{Latex[length=1.8mm]}, semithick, black!70}]
\node[data, text width=3.5cm] (ortho)  at (0,0)   {Orthophotos RGB + CIR\\20\,cm (2013, 2019)};
\node[data, text width=3.5cm] (lidar)  at (4.4,0) {LiDAR nDSM, 1\,m\\(2010/11; 2020 partial)};
\node[data, text width=3.9cm] (ref)    at (9.0,0) {HabitAlp reference polygons\\(2013; 2019/20 re-interpreted)};
\node[proc, text width=3.5cm] (stack)  at (0,-1.6)   {Co-registered 5-band stack\\RGB + NIR + nDSM, 0.5\,m};
\node[proc, text width=4.0cm] (split)  at (4.4,-1.5) {$12\times12$ block split, train/val/test\\(70\,\%/15\,\%/15\,\%); masked split:\\2019/20 test patches + 128\,m buffer\\removed from train/val};
\node[proc, text width=3.9cm] (labels) at (9.0,-1.6) {23 habitat classes (Table~\ref{tab:classes}),\\9 change classes (Table~\ref{tab:transitions})};
\node[box, fill=white, text width=1.5cm] (m1) at (-0.9,-3.75) {MiT-B2 /\\ResNet-50\\U-Net};
\node[box, fill=white, text width=1.5cm] (m2) at (1.1,-3.75)  {Clay v1.0};
\node[box, fill=white, text width=1.5cm] (m3) at (3.1,-3.75)  {Clay v1.5};
\node[box, fill=white, text width=1.5cm] (m4) at (5.1,-3.75)  {Prithvi-\\EO-2.0};
\node[box, fill=white, text width=1.5cm] (m5) at (7.1,-3.75)  {DOFA};
\node[box, fill=white, text width=1.5cm] (m6) at (9.1,-3.75)  {TerraMind};
\node[anchor=north, align=center] (abl) at (4.1,-4.3) {common protocol (Table~\ref{tab:hyperparameters}), three seeds per model; decoder check (one seed);\\input ablation RGB+NIR vs.\ RGB+NIR+nDSM (original runs)};
\begin{scope}[on background layer]
\coordinate (trainpad) at (-0.9,-2.9);
\node[grp, fit=(m1)(m6)(abl)(trainpad)] (train) {};
\end{scope}
\node[glab] at (train.north west) {Fine-tuning on 2013 training blocks};
\node[proc, text width=3.9cm] (inf20) at (6.6,-6.5) {Inference on the 2019/20 test patches\\(2019 imagery + 2020 nDSM)};
\node[data, text width=3.5cm] (map13) at (1.6,-7.7) {2013 reference\\habitat map};
\node[res, text width=3.5cm]  (map20) at (6.6,-7.7) {Predicted habitat map 2019/20};
\node[proc, text width=5.4cm] (cmp)   at (4.1,-9.1) {Pixel-wise comparison with transition rules\\$\rightarrow$ binary and multi-class change maps};
\begin{scope}[on background layer]
\coordinate (cdpad) at (1.6,-5.65);
\node[grp, fit=(map13)(inf20)(cmp)(cdpad)] (cd) {};
\end{scope}
\node[glab] at (cd.north west) {Post-classification change detection};
\node[proc, text width=4.4cm] (pp)   at (1.3,-10.9) {Post-processing (original runs): plausibility\\filter, vectorisation, LiDAR canopy-cover correction};
\node[res, text width=4.4cm]  (eval) at (6.9,-10.9) {Evaluation: macro IoU/F1/BA,\\paired bootstrap 95\% CIs (tiles, seeds)};
\node[res, text width=4.4cm]  (prod) at (1.3,-12.2) {Operational polygon product};
\draw[arr] (ortho) -- (stack);
\draw[arr] (lidar) -- (stack);
\draw[arr] (lidar.south east) -- (labels.north west);
\draw[arr] (ref) -- (labels);
\draw[arr] (stack.south) -- (train.north -| stack.south);
\draw[arr] (split.south) -- (train.north -| split.south);
\draw[arr] (labels.south) -- (train.north -| labels.south);
\draw[arr] (train.south -| inf20.north) -- (inf20.north);
\draw[arr] (inf20) -- (map20);
\draw[arr] (map13.south) -- ++(0,-0.45) -| ([xshift=-8mm]cmp.north);
\draw[arr] (map20.south) -- ++(0,-0.45) -| ([xshift=8mm]cmp.north);
\draw[arr] (cd.south -| pp.north) -- (pp.north);
\draw[arr] (cd.south -| eval.north) -- (eval.north);
\draw[arr] (pp) -- (prod);
\draw[arr] (labels.east) -- ++(0.35,0) coordinate (rl) |- (eval.east);
\coordinate (rly) at (0,-6.6);
\node[font=\tiny, rotate=90, fill=white, inner sep=1.5pt] at (rl |- rly) {2019/20 reference change labels};
\end{tikzpicture}
\caption{Workflow of the post-classification benchmark. Blue: input data; orange: processing steps; green: intermediate and final products. Each model is trained with three seeds (one seed for the decoder check) on the 2013 training blocks of the masked split, from which the 2019/20 test patches plus a 128\,m buffer are removed for training and validation, and applied unchanged to the 2019/20 test epoch. Change maps are derived by comparing each predicted 2019/20 habitat map with the 2013 reference map and are evaluated against the independent 2019/20 reference change labels. The operational post-processing is applied to the original runs (one per model, original split).\label{fig:workflow}}
\end{figure}

\textbf{Band adaptation.} All models receive the same five channels (RGB+NIR+nDSM), but each architecture maps them differently. Clay v1.0 routes the nDSM through the surface-model (DSM) entry of its band registry, so the height channel enters through a pretrained embedding. Clay v1.5 uses NAIP platform metadata (nominal ground sampling distance (GSD) of 1.0\,m) with explicit wavelength encoding for the four spectral bands. DOFA has no height band, so the nDSM is passed through its wavelength-conditioned encoder as a second near-infrared band. Prithvi maps the spectral inputs to its HLS band positions, and TerraMind receives all five bands through its Sentinel-2 L2A modality interface. Neither interface has a height channel, so the nDSM channel of their patch embedding is newly initialised. The MiT-B2 U-Net baseline treats all bands as generic input channels.

\begin{table}[H]
\caption{Post-classification models and key training settings. All use AdamW (wd = 0.05), cosine schedule with warm-up, and RGB+NIR+nDSM inputs in the main configuration. LR = learning rate; HLS = Harmonized Landsat Sentinel-2. Encoder sizes are approximate parameter counts. Patch sizes are in pixels on the 0.5\,m grid (256\,px = 128\,m). All models were trained and evaluated in this study under one protocol; no results are taken from other publications. Decoder check: one seed each.\label{tab:hyperparameters}}
\footnotesize

\begin{adjustwidth}{-\extralength}{0cm}
\begin{tabularx}{\fulllength}{lllllc}
\toprule
\textbf{Model} & \textbf{Backbone/Decoder} & \textbf{Encoder} & \textbf{LR} & \textbf{Pretraining} & \textbf{Patch Size} \\
\midrule
MiT-B2 U-Net & mit\_b2/U-Net & MiT-B2, 25M & 1 $\times$ 10$^{-4}$ %
 & ImageNet & 256 $\times$ 256 \\
Clay v1.0 & clay\_v1/FCN & ViT-B, 86M & 2.5 $\times$ 10$^{-5}$ & Satellite + aerial & 256 $\times$ 256 \\
Clay v1.5 & clay\_v1.5/FCN & ViT-L, 300M & 2.5 $\times$ 10$^{-5}$ & Satellite + aerial & 256 $\times$ 256 \\
Prithvi-EO-2.0 & prithvi\_eo\_v2\_300/FCN & ViT-L, 300M & 5 $\times$ 10$^{-5}$ & HLS (30 m) & 256 $\times$ 256 \\
DOFA & dofa\_base/U-Net & ViT-B, 86M & 1 $\times$ 10$^{-4}$ & Multimodal (incl.\ NAIP) & 256 $\times$ 256 \\
TerraMind & terramind/U-Net & ViT-B, 86M & 6 $\times$ 10$^{-5}$ & Multimodal & 256 $\times$ 256 \\
ResNet-50 U-Net & resnet50/U-Net & ResNet-50, 24M & 1 $\times$ 10$^{-4}$ & ImageNet & 256 $\times$ 256 \\
\midrule
\multicolumn{6}{l}{\textit{Decoder check %
}} \\
Clay v1.0 & clay\_v1/U-Net & ViT-B, 86M & 2.5 $\times$ 10$^{-5}$ & Satellite + aerial & 256 $\times$ 256 \\
Prithvi-EO-2.0 & prithvi\_eo\_v2\_300/U-Net & ViT-L, 300M & 5 $\times$ 10$^{-5}$ & HLS (30 m) & 256 $\times$ 256 \\
\bottomrule
\end{tabularx}
\end{adjustwidth}
\end{table}

\textbf{Change detection.} Each model is trained on 2013 reference data and applied to the independent 2019/20 test epoch to produce a predicted 2019/20 habitat map. A rule-based, per-pixel comparison of the 2013 reference map with this prediction assigns each pixel to one of the eight transition classes or ``no change'' (Table~\ref{tab:transitions}). The reference change map is derived with the same rules from the 2013 and 2019/20 reference maps. This mirrors the operational setting, in which an existing expert map is updated from new imagery, and it means that change errors stem from the 2019/20 prediction alone rather than from two predicted maps. The 2019/20 test patches lie within the 2013 training extent, since change detection requires predicting the same locations at both dates, but the test-epoch imagery, LiDAR, and labels are never seen during training. Inference uses 256$\times$256 patches for all models with 50\% overlap. An 8-pixel border of each patch is discarded, the logits of overlapping patches are averaged with weights that fall from 1 to 0 over the outer 64 pixels along a cosine ramp, and each pixel receives the class with the highest averaged logit.

\textbf{LiDAR ablation.} To quantify the value of LiDAR (RQ3), each model is trained with RGB+NIR and with RGB+NIR+nDSM, and we report the resulting change-detection difference. For single-date segmentation we additionally compare RGB+NIR against RGB+NIR+nDSM. Single-date segmentation is evaluated on the 2013 in-domain test blocks and on the 2019/20 test patches. Because the reference derives the canopy-cover levels \mbox{CC $<$ 80} and CC $\geq$ 80 from LiDAR, four of the eight change classes (149 of 476\,ha of reference change) depend on a LiDAR threshold, so we also evaluate both tasks with each\linebreak CC $<$ 80/CC $\geq$ 80 distinction removed. %

\textbf{Metrics.} Because ``no change'' dominates (58.9\% of the evaluated 2013 to 2019/20~reference), we use \emph{macro IoU} as the primary metric and report \emph{macro F1} alongside it, both computed with equal weight over the nine classes (or two, for the binary change/no-change task). We additionally report \emph{balanced accuracy} (BA; the mean of per-class recall, i.e.,\ macro-averaged accuracy) as a class-balanced summary, and we avoid micro overall accuracy, which is inflated by the majority class. Following the reporting standard of Corley et~al. \cite{Corley2026SOTA}, every headline value is the mean $\pm$ standard deviation over the three training seeds, and a difference between two models is called robust only when its 95\% confidence interval (CI) excludes zero. CIs of differences come from a paired hierarchical bootstrap with 10{,}000~replicates: in each replicate the 88 test tiles (500\,m $\times$ 500\,m, clipped to the test patches) and, per model, the three seeds are resampled with replacement, and the difference in seed-mean macro IoU is computed on the same tiles for both models. Tiles rather than pixels are resampled because neighbouring pixels are spatially correlated \cite{Lyons2018Comparison}, and a replicate that happens to miss a class is drawn again, so macro IoU always averages over the full class set. The percentile interval thus covers both test-area sampling and training randomness. We report unadjusted intervals and no p-values. The comparison of Clay v1.0 with Prithvi-EO-2.0 was added after the first submission and is marked as post hoc. For the original single runs (LiDAR ablation, post-processing), the tile bootstrap alone (1000 replicates) gives CIs that cover test-area sampling only.

\subsection{Operational Post-Processing}
\label{sec:pp_methods}

To move from pixel predictions toward an operational product (RQ4), we apply a rule-based post-processing pipeline (applied on top of the model output, never used during training): (i) \emph{physical-plausibility filtering} removes six types of ecologically implausible transitions (non-forest\,$\rightarrow$\,mature or old forest, young growth\,$\rightarrow$\,old forest, regression to a younger forest stage, rock\,$\rightarrow$\,grassland, water on slopes steeper than 35$^\circ$, and alpine vegetation below 650\,m; slope and elevation from the digital terrain model of GIS Steiermark, %
 \url{https://gis.stmk.gv.at/image/rest/services/OGD_Hoehen}, accessed on 17 November 2026) by resetting violating pixels to their 2013 reference class so that they count as ``no change''; (ii) \emph{vectorisation} onto the existing HabitAlp polygon geometry; and (iii) a \emph{LiDAR-based canopy-cover update}, which assigns a forest polygon the CC $\geq$ 80 level of its class when at least 80\% of its area has an nDSM above 1.3\,m, and the CC $<$ 80 level otherwise. In step (ii), a 2013 polygon receives the majority class of the filtered map from step (i) if this class covers more than 90\% of its area (including pixels without a prediction), or more than 60\% when the rest is smaller than 0.1\,ha. Polygons whose majority class falls below these thresholds but covers more than 30\% are split along the regions of the map smoothed with a $31\times31$ pixel (6.2\,m) mode filter. Every smoothed region larger than 0.1\,ha becomes a separate part, except the largest, which remains the rest of the polygon; the part outlines are then simplified (Visvalingam--Whyatt, area threshold 1.24\,m$^2$), parts smaller than 0.1\,ha are merged into their largest adjacent part, and each part receives its majority class in the unsmoothed map. If the majority class covers 30\% or less, the polygon stays unlabelled; this affects about 1\% of the test area, which is excluded from the post-processed metrics. Steps (i) and (ii) draw on the 2013 reference map and its polygon delineation. These are legitimate operational inputs, since the previous expert map is always available when a map is updated, but they are not model output, so metrics after post-processing no longer measure the models alone. We therefore use the model-only (no post-processing) metrics for the benchmark (RQ1--RQ3) and report post-processing separately to answer RQ4. Steps (ii) and (iii) were run in IMPACT Tools, proprietary software of JOANNEUM RESEARCH for in-house use only, so instead of code we release their outputs (polygons and rasters) together with the model predictions with the dataset, from which the post-processing results can be recomputed. The operational post-processing was applied to the original runs (one per model, original split). The original Clay v1.0 run lies at the upper end of its three revision seeds (multi-class change macro IoU 0.240 vs. 0.230 to 0.235), so its post-processed result is representative of the model, whereas the original MiT-B2 U-Net run (batch size 8) scores above its revision seeds (2019/20 segmentation 0.265 vs. 0.214 to 0.242).

\textbf{Use of generative AI.} During the preparation of this manuscript, the authors used Claude Opus 4.6 and, for the revision, Claude Opus 5.5 and Claude Fable 5.1 (Anthropic, San Francisco, CA, USA) for the purposes of drafting support, analysis scripting, and language editing. The authors have reviewed and edited the output and take full responsibility for the content of this publication.

\section{Results}
\label{sec:results}

\subsection{Single-Date Semantic Segmentation}

Single-date habitat segmentation, the basis for the change maps, is itself challenging (Table~\ref{tab:r1-b9_main_seg}). On the 2013 in-domain test blocks, all seven models reach between 0.285 and 0.329 macro IoU, with Clay v1.0 highest. On the 2019/20 test patches every model loses 0.08 to 0.13, and Clay v1.0 loses least and stays highest, ahead of the MiT-B2 U-Net by +0.023 (95\% CI [+0.006, +0.040]). %
The other four GFMs score below the MiT-B2 U-Net, but the difference is robust only for DOFA, borderline for TerraMind (upper CI bound just below zero, $-$0.0004), and not robust for Clay v1.5 and Prithvi-EO-2.0. The ResNet-50 U-Net is lowest (0.157). The seed standard deviation is largest for the two U-Nets and Prithvi-EO-2.0 (0.013 to 0.015) and at most 0.006 for the other GFMs.

Per-class results for 2019/20 (Table~\ref{tab:r1-b9_seg_perclass} in Appendix~\ref{app:perclass}) show the same structure for every model. Mountain dwarf forest has the highest or tied highest IoU for six of the seven models (0.36 to 0.47): it forms large, contiguous stands in the study area and is the second-largest class in the training labels (16.5\%). The lowest scores, for gravel banks and the CC $<$ 80 levels of broad-leaved mature and old forest (at most 0.15), belong to the three smallest training classes (each below 0.6\% of the labels). Clay v1.0 is best or tied best in 14 of the 23 classes.

\begin{table}[H]

\caption{Single-date semantic segmentation (23 classes, RGB+NIR+nDSM) on the 2013 in-domain test blocks and the 2019/20 cross-temporal test patches: mean $\pm$ standard deviation over three training seeds (revision runs: masked split, best validation checkpoint; all models trained and evaluated in this study under one protocol, no values taken from other publications). BA = balanced accuracy; IoU and F1 are macro averages; all at 0.2\,m. Best mean per column in bold, from unrounded values.\label{tab:r1-b9_main_seg}}
\begin{adjustwidth}{-\extralength}{0cm}
\begin{tabularx}{\fulllength}{lCCCCCC}
\toprule
& \multicolumn{3}{c}{\textbf{2013 (in-Domain)}} &  \multicolumn{3}{c}{\textbf{2019/20 (Cross-Temporal)}} \\
\cmidrule{2-7}
\textbf{Model} & \textbf{BA} & \textbf{IoU} & \textbf{F1} &  \textbf{BA} & \textbf{IoU} & \textbf{F1} \\
\midrule
Clay v1.0 & \textbf{0.492} $\pm$ 0.005 & \textbf{0.329} $\pm$ 0.003 & \textbf{0.472} $\pm$ 0.004 &   \textbf{0.412} $\pm$ 0.004 & \textbf{0.249} $\pm$ 0.003 & \textbf{0.389} $\pm$ 0.004 \\
Clay v1.5 & 0.476 $\pm$ 0.003 & 0.311 $\pm$ 0.002 & 0.452 $\pm$ 0.003 &   0.351 $\pm$ 0.008 & 0.210 $\pm$ 0.006 & 0.338 $\pm$ 0.009 \\
Prithvi-EO-2.0 & 0.470 $\pm$ 0.006 & 0.304 $\pm$ 0.013 & 0.446 $\pm$ 0.015 &   0.358 $\pm$ 0.008 & 0.216 $\pm$ 0.013 & 0.348 $\pm$ 0.016 \\
DOFA & 0.471 $\pm$ 0.004 & 0.294 $\pm$ 0.003 & 0.435 $\pm$ 0.003 &   0.354 $\pm$ 0.008 & 0.201 $\pm$ 0.004 & 0.327 $\pm$ 0.006 \\
TerraMind & 0.468 $\pm$ 0.006 & 0.306 $\pm$ 0.005 & 0.447 $\pm$ 0.005 &   0.349 $\pm$ 0.009 & 0.208 $\pm$ 0.003 & 0.336 $\pm$ 0.004 \\
MiT-B2 U-Net & 0.488 $\pm$ 0.008 & 0.312 $\pm$ 0.009 & 0.452 $\pm$ 0.013 &   0.400 $\pm$ 0.020 & 0.226 $\pm$ 0.015 & 0.359 $\pm$ 0.021 \\
ResNet-50 U-Net & 0.463 $\pm$ 0.017 & 0.285 $\pm$ 0.010 & 0.423 $\pm$ 0.013 &   0.283 $\pm$ 0.027 & 0.157 $\pm$ 0.014 & 0.265 $\pm$ 0.021 \\
\bottomrule
\end{tabularx}
\end{adjustwidth}
\end{table}

\subsection{Cross-Temporal Change Detection: GFMs vs. Baselines}

\begin{figure}[H]
    \includegraphics[width=\textwidth]{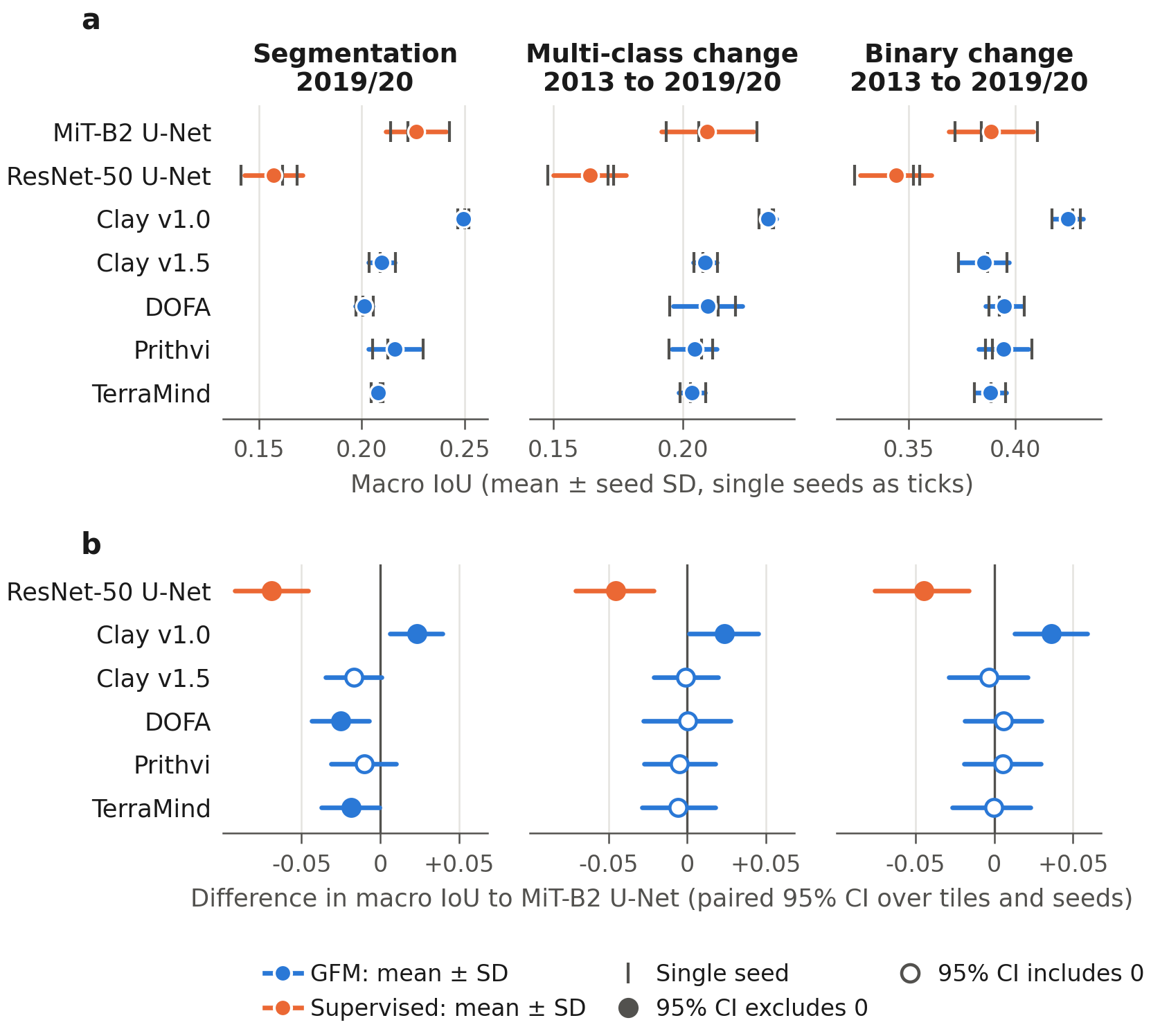}
    \caption{Model comparison on the 2019/20 test patches (revision runs: masked split, best-validation checkpoint, three training seeds per model). (\textbf{a}) Macro IoU for 2019/20 segmentation and for multi-class and binary change detection (2013 to 2019/20): seed mean $\pm$ seed standard deviation, single seeds as ticks. (\textbf{b}) Difference in macro IoU to the MiT-B2 U-Net with 95\% CIs from a paired hierarchical bootstrap over the 88 test tiles (500\,m) and the three seeds (10{,}000 replicates); filled markers: the interval excludes zero. Blue: GFMs; orange: supervised baselines.\label{fig:benchmark}}
\end{figure}

Table~\ref{tab:r1-b9_main_cd} and Figure~\ref{fig:benchmark} report the central result, cross-temporal change detection from 2013 to 2019/20. Clay v1.0 has the highest mean in all three metrics for binary and multi-class change. It is also the only GFM ahead of the MiT-B2 U-Net, robustly for binary change and only just for multi-class change, where the lower CI bound is +0.001 (Table~\ref{tab:r1-b1_paired}). This margin (+0.024) is small, its lower CI bound is close to zero, and the intervals are not adjusted for multiple comparisons, so it is indicative rather than established. Clay v1.5, Prithvi-EO-2.0, DOFA, and TerraMind are level with the MiT-B2 U-Net: they lie within 0.006~macro IoU of it in both tasks, and none of these differences is robust. The convolutional ResNet-50 U-Net is robustly below the MiT-B2 U-Net and, in post hoc comparisons, below every GFM in all three metrics (by 0.040 to 0.069 multi-class macro IoU). Clay v1.0 is also robustly ahead of Prithvi-EO-2.0 in all three metrics (post hoc). Balanced accuracy follows the same ordering but sits well above IoU and F1, as expected for an imbalanced problem, and binary balanced accuracy stays only modestly above the 0.50 chance level.
\begin{table}[H]

\caption{Cross-temporal change detection (2013$\rightarrow$2019/20, RGB+NIR+nDSM, model output before post-processing), binary and nine classes: mean $\pm$ standard deviation over three training seeds (revision runs: masked split, best validation checkpoint; all models trained and evaluated in this study under one protocol, no values taken from other publications). BA = balanced accuracy; IoU and F1 are macro averages; all at 0.2\,m. Best mean per column in bold, from unrounded values. Paired differences to the MiT-B2 U-Net with 95\% CIs: Table~\ref{tab:r1-b1_paired} and Figure~\ref{fig:benchmark}.\label{tab:r1-b9_main_cd}}
\begin{adjustwidth}{-\extralength}{0cm}
\begin{tabularx}{\fulllength}{lCCCCCC}
\toprule
& \multicolumn{3}{c}{\textbf{Binary}} &  \multicolumn{3}{c}{\textbf{Multi-Class (9)}} \\
\cmidrule{2-7}
\textbf{Model} & \textbf{BA} & \textbf{IoU} & \textbf{F1} &  \textbf{BA} &\textbf{IoU} & \textbf{F1} \\
\midrule
Clay v1.0 & \textbf{0.633} $\pm$ 0.007 & \textbf{0.425} $\pm$ 0.007 & \textbf{0.596} $\pm$ 0.007 &   \textbf{0.487} $\pm$ 0.002 & \textbf{0.233} $\pm$ 0.003 & \textbf{0.362} $\pm$ 0.004 \\
Clay v1.5 & 0.603 $\pm$ 0.012 & 0.386 $\pm$ 0.011 & 0.554 $\pm$ 0.012 &   0.464 $\pm$ 0.013 & 0.209 $\pm$ 0.005 & 0.331 $\pm$ 0.006 \\
Prithvi-EO-2.0 & 0.605 $\pm$ 0.011 & 0.394 $\pm$ 0.012 & 0.564 $\pm$ 0.012 &   0.451 $\pm$ 0.013 & 0.204 $\pm$ 0.009 & 0.326 $\pm$ 0.012 \\
DOFA & 0.613 $\pm$ 0.008 & 0.395 $\pm$ 0.009 & 0.564 $\pm$ 0.009 &   0.459 $\pm$ 0.017 & 0.210 $\pm$ 0.013 & 0.334 $\pm$ 0.017 \\
TerraMind & 0.598 $\pm$ 0.007 & 0.388 $\pm$ 0.007 & 0.558 $\pm$ 0.008 &   0.461 $\pm$ 0.005 & 0.204 $\pm$ 0.005 & 0.326 $\pm$ 0.008 \\
MiT-B2 U-Net & 0.595 $\pm$ 0.018 & 0.389 $\pm$ 0.020 & 0.559 $\pm$ 0.021 &   0.456 $\pm$ 0.016 & 0.209 $\pm$ 0.018 & 0.333 $\pm$ 0.023 \\
ResNet-50 U-Net & 0.564 $\pm$ 0.017 & 0.344 $\pm$ 0.017 & 0.508 $\pm$ 0.019 &   0.403 $\pm$ 0.016 & 0.164 $\pm$ 0.014 & 0.273 $\pm$ 0.021 \\
\bottomrule
\end{tabularx}
\end{adjustwidth}
\end{table}
\vspace{-6pt}

Two checks qualify this ranking (Table~\ref{tab:r1-b1_paired}, supporting comparisons; Table~\ref{tab:r1-b9_seed_summary}). With a U-Net decoder for both models (one seed each), Clay v1.0 stays robustly ahead of Prithvi-EO-2.0 in segmentation, but not in change detection, because the U-Net decoder lowers Clay v1.0 slightly and raises Prithvi-EO-2.0 in multi-class change. For the MiT-B2 U-Net, the masked, original, and block-drop splits give segmentation scores within 0.013 of each other, and the only robust difference in change detection is a higher multi-class score on the masked than on the original split. Clay v1.0's lead comes at a cost: it needs about 45 times the computation of the MiT-B2 U-Net per patch (Table~\ref{tab:r1-b10_efficiency}).

\begin{table}[H]

\caption{Paired differences in macro IoU (0.2\,m) between models, mean over seeds (best checkpoint), with unadjusted 95\% CI from a paired hierarchical bootstrap (10,000 replicates): in each replicate, the 88 test tiles (500\,m) and, per model, its three seeds are resampled with replacement, so the interval covers test-area sampling and training randomness. Bold: the 95\% CI excludes zero. $^\dagger$ Post hoc. U-Net decoder pair: one seed each, tiles only.\label{tab:r1-b1_paired}}
\tablesize{\footnotesize} 
\begin{adjustwidth}{-\extralength}{0cm}
\begin{tabularx}{\fulllength}{Lrlrlrl}
\toprule
\textbf{Comparison} & \multicolumn{2}{c}{\textbf{Seg.\ 2019/20}} & \multicolumn{2}{c}{\textbf{CD Binary}} & \multicolumn{2}{c}{\textbf{CD Multi-Class}} \\
\midrule
Clay v1.0 $-$ MiT-B2 U-Net & \textbf{+0.023} & \textbf{[+0.006, +0.040]} & \textbf{+0.036} & \textbf{[+0.013, +0.059]} & \textbf{+0.024} & \textbf{[+0.001, +0.045]} \\
Clay v1.5 $-$ MiT-B2 U-Net & $-$0.017 & [$-$0.035, +0.001] & $-$0.003 & [$-$0.029, +0.021] & $-$0.001 & [$-$0.021, +0.020] \\
Prithvi-EO-2.0 $-$ MiT-B2 U-Net & $-$0.010 & [$-$0.031, +0.010] & +0.006 & [$-$0.019, +0.030] & $-$0.005 & [$-$0.027, +0.018] \\
DOFA $-$ MiT-B2 U-Net & \textbf{$-$0.025} & \textbf{[$-$0.044, $-$0.007]} & +0.006 & [$-$0.019, +0.030] & 0.000 & [$-$0.028, +0.028] \\
TerraMind $-$ MiT-B2 U-Net & \textbf{$-$0.018} & \textbf{[$-$0.037, $-$0.000]} & 0.000 & [$-$0.026, +0.023] & $-$0.006 & [$-$0.029, +0.018] \\
ResNet-50 U-Net $-$ MiT-B2 U-Net & \textbf{$-$0.069} & \textbf{[$-$0.093, $-$0.046]} & \textbf{$-$0.045} & \textbf{[$-$0.076, $-$0.016]} & \textbf{$-$0.045} & \textbf{[$-$0.071, $-$0.021]} \\
Clay v1.0 $-$ Prithvi-EO-2.0 $^\dagger$ & \textbf{+0.033} & \textbf{[+0.018, +0.048]} & \textbf{+0.031} & \textbf{[+0.011, +0.049]} & \textbf{+0.029} & \textbf{[+0.011, +0.045]} \\
\midrule
\multicolumn{7}{l}{\textit{Against the ResNet-50 U-Net}} \\
Clay v1.0 $-$ ResNet-50 U-Net $^\dagger$ & \textbf{+0.092} & \textbf{[+0.072, +0.112]} & \textbf{+0.081} & \textbf{[+0.056, +0.108]} & \textbf{+0.069} & \textbf{[+0.049, +0.090]} \\
Clay v1.5 $-$ ResNet-50 U-Net $^\dagger$ & \textbf{+0.052} & \textbf{[+0.034, +0.070]} & \textbf{+0.042} & \textbf{[+0.017, +0.066]} & \textbf{+0.044} & \textbf{[+0.026, +0.065]} \\
Prithvi-EO-2.0 $-$ ResNet-50 U-Net $^\dagger$ & \textbf{+0.059} & \textbf{[+0.037, +0.079]} & \textbf{+0.050} & \textbf{[+0.027, +0.075]} & \textbf{+0.040} & \textbf{[+0.020, +0.062]} \\
DOFA $-$ ResNet-50 U-Net $^\dagger$ & \textbf{+0.044} & \textbf{[+0.025, +0.062]} & \textbf{+0.051} & \textbf{[+0.030, +0.073]} & \textbf{+0.046} & \textbf{[+0.023, +0.069]} \\
TerraMind $-$ ResNet-50 U-Net $^\dagger$ & \textbf{+0.051} & \textbf{[+0.034, +0.066]} & \textbf{+0.044} & \textbf{[+0.021, +0.068]} & \textbf{+0.040} & \textbf{[+0.020, +0.061]} \\
\midrule
\multicolumn{7}{l}{\textit{Supporting comparisons}} \\
Clay v1.0, U-Net decoder $-$ Prithvi-EO-2.0, U-Net decoder & \textbf{+0.022} & \textbf{[+0.009, +0.035]} & +0.012 & [$-$0.007, +0.031] & +0.006 & [$-$0.009, +0.024] \\
MiT-B2 U-Net $-$ MiT-B2 U-Net, original split & $-$0.001 & [$-$0.017, +0.016] & +0.004 & [$-$0.019, +0.029] & \textbf{+0.026} & \textbf{[+0.001, +0.051]} \\
MiT-B2 U-Net $-$ MiT-B2 U-Net, block-drop split & +0.012 & [$-$0.004, +0.028] & +0.011 & [$-$0.010, +0.033] & +0.006 & [$-$0.016, +0.030] \\
MiT-B2 U-Net, original split $-$ MiT-B2 U-Net, block-drop split & \textbf{+0.013} & \textbf{[+0.001, +0.025]} & +0.007 & [$-$0.010, +0.024] & $-$0.020 & [$-$0.040, +0.002] \\
\bottomrule
\end{tabularx}
\end{adjustwidth}
\end{table}

\subsection{Per-Class Detectability and Error Structure}

\begin{figure}[H]
   
\begin{adjustwidth}{-\extralength}{0cm}
 \centering
    \includegraphics[width=\textwidth]{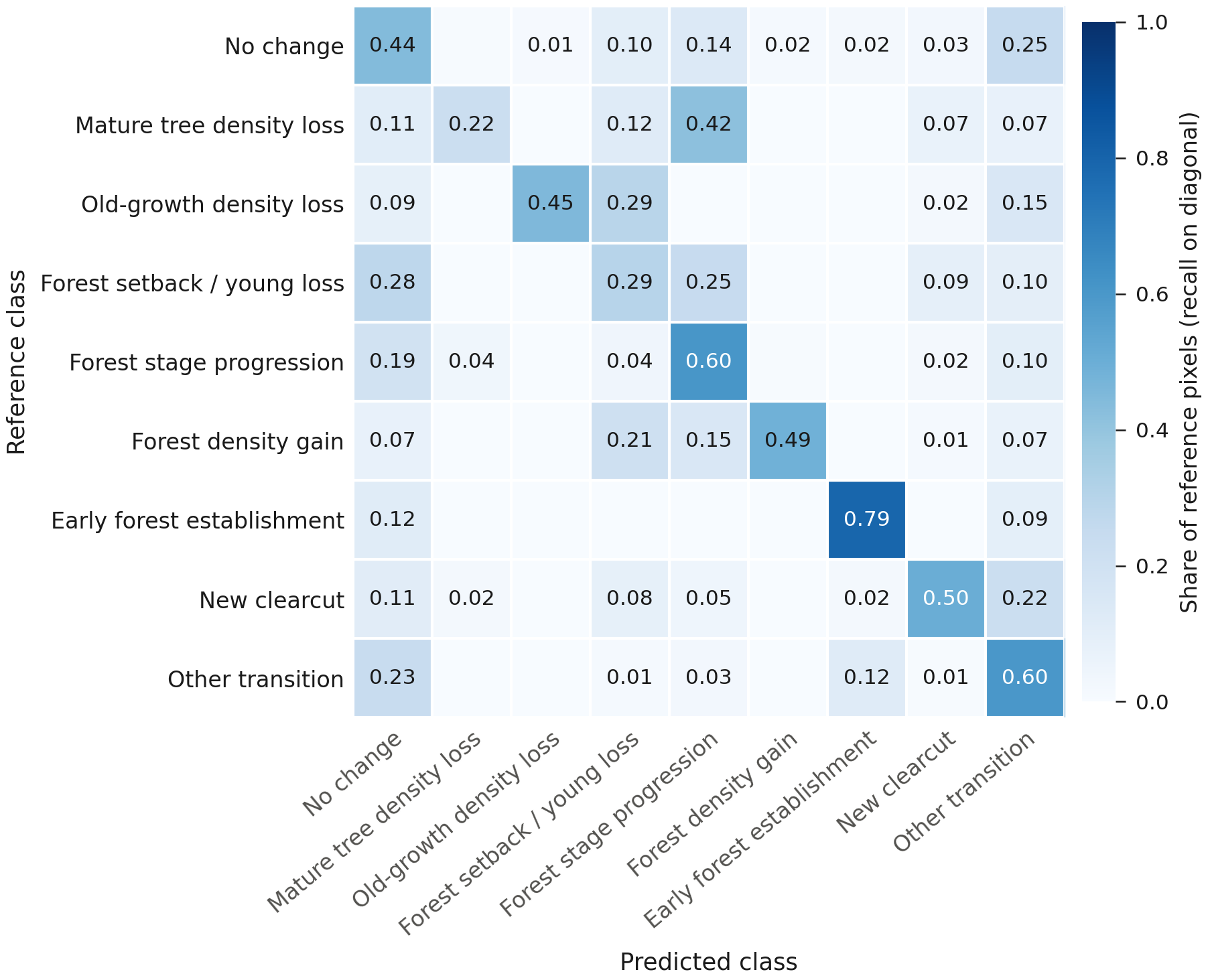}
\end{adjustwidth}
    \caption{Row-normalised confusion matrix over the nine change classes (including no change, 2013 to 2019/20) for Clay v1.0 (revision runs, masked split, pooled over three training seeds). Diagonal values are per-class recall.\label{fig:confusion}}
\end{figure}

Detectability depends on both class frequency and ecological distinctness (Table~\ref{tab:r1-b9_clay_perclass}; Figure~\ref{fig:confusion}, for Clay v1.0). The two abundant classes, no change (58.9\% of the area) and forest stage progression (22.8\%), and the structurally distinct old-growth density loss reach IoU 0.38 to 0.40. Frequency alone does not explain detectability: mature tree density loss is the third-largest class (7.5\%) but reaches a recall of only 0.22, because 42\% of its pixels are predicted as stage progression. Forest setback/young loss and other transition, a heterogeneous residual class, stay below 0.10 IoU, and several rare classes have high recall but low precision (e.g.,\ early forest establishment), i.e.,\ the model over-predicts them. Besides confusions between adjacent forest-development states, the confusion matrix shows an exchange with no change in both directions: no change absorbs about a quarter of forest setback and other transition, while a quarter of the no-change area is predicted as other transition, which largely explains the low precision of that class. Per-class IoU for all seven models (Table~\ref{tab:r1-b9_cd_perclass} in Appendix~\ref{app:perclass}) shows that the ranking is not uniform across classes: Clay v1.0 leads on no change, stage progression and old-growth density loss, but the MiT-B2 U-Net is best on new clearcut, TerraMind on mature tree density loss and early forest establishment, and DOFA on forest density gain.

\begin{table}[H]
\caption{Per-class change-detection metrics for Clay v1.0 (2013$\rightarrow$2019/20, model output before post-processing; revision runs, masked split, mean over three training seeds). Area: share of the evaluated reference change map.\label{tab:r1-b9_clay_perclass}}
\begin{tabularx}{\textwidth}{lrCCCC}
\toprule
\textbf{Transition Class} & \textbf{Area (\%)} & \textbf{IoU} & \textbf{F1} & \textbf{Recall} & \textbf{Precision} \\
\midrule
No change & 58.9 & 0.39 & 0.56 & 0.44 & 0.79 \\
Forest stage progression & 22.8 & 0.40 & 0.57 & 0.60 & 0.54 \\
Old-growth density loss & 2.8 & 0.38 & 0.55 & 0.45 & 0.71 \\
Early forest establishment & 0.8 & 0.25 & 0.40 & 0.79 & 0.27 \\
Mature tree density loss & 7.5 & 0.19 & 0.32 & 0.22 & 0.60 \\
New clearcut & 1.5 & 0.18 & 0.30 & 0.50 & 0.21 \\
Forest density gain & 0.6 & 0.17 & 0.30 & 0.49 & 0.21 \\
Other transition & 3.1 & 0.09 & 0.16 & 0.60 & 0.09 \\
Forest setback/young loss & 1.9 & 0.05 & 0.10 & 0.29 & 0.06 \\
\midrule
Macro average &  & 0.23 & 0.36 & 0.49 & 0.39 \\
\bottomrule
\end{tabularx}
\end{table}
\vspace{-6pt}

\subsection{Value of LiDAR Height Information}

Adding the LiDAR-derived nDSM to RGB+NIR improves the change detection of the model output (original runs, one per model, original split; Table~\ref{tab:r1-b3_cd_ablation}): averaged over the six models, macro IoU rises by +0.041 for binary and +0.050 for multi-class change, and the multi-class gain is robust for every model. Because these are single runs, the CIs cover test-area sampling only. Most of this gain sits in the four change classes whose reference is defined by the LiDAR canopy-cover threshold: with each CC $<$ 80/CC $\geq$ 80 distinction removed, the mean gain shrinks to +0.011 and is robust only for DOFA. After the operational post-processing, which itself updates canopy cover from the LiDAR, the mean gain is close to zero in all three settings.

\startlandscape
\begin{table}[H]

\caption{Effect of adding the LiDAR nDSM on cross-temporal change detection %
(2013$\rightarrow$2019/20, original runs, one per model): macro IoU of RGB+NIR and RGB+NIR+nDSM and $\Delta$ from unrounded values, for the model output and after post-processing. CC collapsed: each CC $<$ 80/CC $\geq$ 80 distinction removed, so the four canopy-cover change classes merge into no change or other transition (four change classes plus no change). 95\% CI of $\Delta$: paired spatial block bootstrap over 500\,m tiles, 1000 resamples (88 tiles); test-set sampling only, the original runs are single runs.\label{tab:r1-b3_cd_ablation}}
\begin{tabularx}{\textwidth}{lCCClCCClCCCl}
\toprule
& \multicolumn{4}{c}{\textbf{Binary CD}} & \multicolumn{4}{c}{\textbf{Multi-Class CD (9 Classes)}} & \multicolumn{4}{c}{\textbf{Multi-Class CD, CC Collapsed (5)}} \\
\cmidrule{2-13}
\textbf{Model} & \textbf{RGB+NIR} & \textbf{+nDSM} &\boldmath{$\Delta$} & \textbf{[95\% CI]} & \textbf{RGB+NIR} & \textbf{+nDSM} & \boldmath{$\Delta$} & \textbf{[95\% CI]} & \textbf{RGB+NIR} & \textbf{+nDSM} & \boldmath{$\Delta$} & \textbf{[95\% CI]} \\
\midrule
\multicolumn{13}{l}{\textit{Model output}} \\
Clay v1.0 & 0.375 & 0.431 & +0.055 & [+0.030, +0.082] & 0.177 & 0.240 & +0.063 & [+0.040, +0.084] & 0.241 & 0.259 & +0.018 & [$-$0.002, +0.037] \\
Clay v1.5 & 0.380 & 0.396 & +0.015 & [$-$0.006, +0.039] & 0.163 & 0.207 & +0.044 & [+0.021, +0.065] & 0.239 & 0.226 & $-$0.013 & [$-$0.031, +0.009] \\
Prithvi-EO-2.0 & 0.351 & 0.362 & +0.011 & [$-$0.016, +0.045] & 0.152 & 0.173 & +0.020 & [+0.004, +0.038] & 0.218 & 0.214 & $-$0.005 & [$-$0.027, +0.020] \\
DOFA & 0.315 & 0.406 & +0.091 & [+0.063, +0.120] & 0.128 & 0.207 & +0.080 & [+0.058, +0.101] & 0.181 & 0.236 & +0.055 & [+0.028, +0.078] \\
TerraMind & 0.327 & 0.369 & +0.042 & [+0.017, +0.069] & 0.146 & 0.196 & +0.050 & [+0.029, +0.073] & 0.181 & 0.204 & +0.023 & [$-$0.001, +0.046] \\
MiT-B2 U-Net & 0.392 & 0.420 & +0.028 & [+0.006, +0.050] & 0.168 & 0.212 & +0.044 & [+0.017, +0.066] & 0.253 & 0.239 & $-$0.014 & [$-$0.035, +0.005] \\
\midrule
Mean (6 models) & 0.357 & 0.397 & +0.041 & - & 0.156 & 0.206 & +0.050 & - & 0.219 & 0.230 & +0.011 & - \\
\midrule
\multicolumn{13}{l}{\textit{After post-processing}} \\
Clay v1.0 & 0.428 & 0.448 & +0.019 & [$-$0.009, +0.049] & 0.235 & 0.207 & $-$0.028 & [$-$0.053, $-$0.004] & 0.313 & 0.316 & +0.003 & [$-$0.030, +0.035] \\
Clay v1.5 & 0.439 & 0.426 & $-$0.013 & [$-$0.038, +0.012] & 0.204 & 0.209 & +0.005 & [$-$0.017, +0.026] & 0.303 & 0.302 & 0.000 & [$-$0.026, +0.027] \\
Prithvi-EO-2.0 & 0.413 & 0.401 & $-$0.012 & [$-$0.044, +0.023] & 0.220 & 0.210 & $-$0.009 & [$-$0.041, +0.024] & 0.293 & 0.279 & $-$0.014 & [$-$0.044, +0.023] \\
DOFA & 0.371 & 0.439 & +0.069 & [+0.037, +0.099] & 0.205 & 0.220 & +0.015 & [$-$0.010, +0.038] & 0.258 & 0.307 & +0.050 & [+0.016, +0.078] \\
TerraMind & 0.386 & 0.389 & +0.003 & [$-$0.028, +0.033] & 0.212 & 0.187 & $-$0.025 & [$-$0.049, +0.001] & 0.266 & 0.256 & $-$0.010 & [$-$0.043, +0.021] \\
MiT-B2 U-Net & 0.454 & 0.424 & $-$0.030 & [$-$0.050, $-$0.010] & 0.224 & 0.211 & $-$0.014 & [$-$0.036, +0.007] & 0.320 & 0.306 & $-$0.013 & [$-$0.046, +0.015] \\
\midrule
Mean (6 models) & 0.415 & 0.421 & +0.006 & - & 0.217 & 0.207 & $-$0.009 & - & 0.292 & 0.295 & +0.002 & - \\
\bottomrule
\end{tabularx}
\end{table}
\finishlandscape

Single-date segmentation shows the same pattern (Table~\ref{tab:r1-b4_ndsm_ablation_mean}; per model in Table~\ref{tab:r1-b4_ndsm_ablation}). On the 2019/20 test patches, the six-model mean over all 23 classes changes by $-0.009$, which is not robust: the 10 canopy-cover classes gain 0.052, robustly and for every model, while the other 13 classes lose 0.056, mostly in the non-forest classes. With each canopy-cover pair merged into one stage class, the gain shrinks to +0.017 and is no longer robust. On the 2013 test blocks the mean effect is about zero.

\begin{table}[H]

\caption{Effect of the LiDAR nDSM on single-date segmentation, mean over the six models: macro IoU of RGB+NIR+nDSM (E) and RGB+NIR (F) and $\Delta$ = E $-$ F by class group, at 0.2\,m. 95\% CI of $\Delta$: paired spatial block bootstrap over 500\,m tiles, 1000 resamples, all models on the same resamples.\label{tab:r1-b4_ndsm_ablation_mean}}

\begin{adjustwidth}{-\extralength}{0cm}
\begin{tabularx}{\fulllength}{lCCCcCCCc}
\toprule
& \multicolumn{4}{c}{\textbf{2019/20 (Cross-Temporal)}} & \multicolumn{4}{c}{\textbf{2013 (in-Domain)}} \\
\cmidrule{2-9}
\textbf{Class Group} & \textbf{E} & \textbf{F} & \boldmath{$\Delta$} & \textbf{[95\% CI]} & \textbf{E} & \textbf{F} & \boldmath{$\Delta$} & \textbf{[95\% CI]} \\
\midrule
All 23 classes & 0.220 & 0.229 & $-$0.009 & [$-$0.024, +0.004] &   0.300 & 0.296 & +0.003 & [$-$0.002, +0.009] \\
Canopy-cover classes (10) & 0.186 & 0.134 & +0.052 & [+0.037, +0.067] &  0.199 & 0.191 & +0.008 & [0.000, +0.016] \\
Other classes (13) & 0.246 & 0.303 & $-$0.056 & [$-$0.075, $-$0.040] & 0.377 & 0.377 & 0.000 & [$-$0.007, +0.007] \\
\quad forest without CC split (5) & 0.254 & 0.251 & +0.003 & [$-$0.022, +0.024] &  0.353 & 0.334 & +0.019 & [+0.006, +0.032] \\
\quad non-forest (8) & 0.242 & 0.335 & $-$0.093 & [$-$0.115, $-$0.073] & 0.393 & 0.405 & $-$0.012 & [$-$0.021, $-$0.003] \\
\midrule
Canopy-cover pairs merged (5) & 0.266 & 0.249 & +0.017 & [$-$0.005, +0.040] &  0.277 & 0.275 & +0.001 & [$-$0.012, +0.015] \\
\bottomrule
\end{tabularx}
\end{adjustwidth}
\noindent{\footnotesize{Canopy-cover classes: 8, 9, 11--18 (pole, mature and old forest split at 80\% LiDAR canopy cover). Forest without CC split: 6, 7, 10, 19, 20. Non-forest: 1--5, 21--23. Merged: each CC $<$ 80/CC $\geq$ 80 pair counted as one stage class.}} %

\end{table}

\subsection{From Pixels to an Operational Product}
\label{sec:pp_results}

\begin{figure}[H]
    \includegraphics[width=0.99\textwidth]{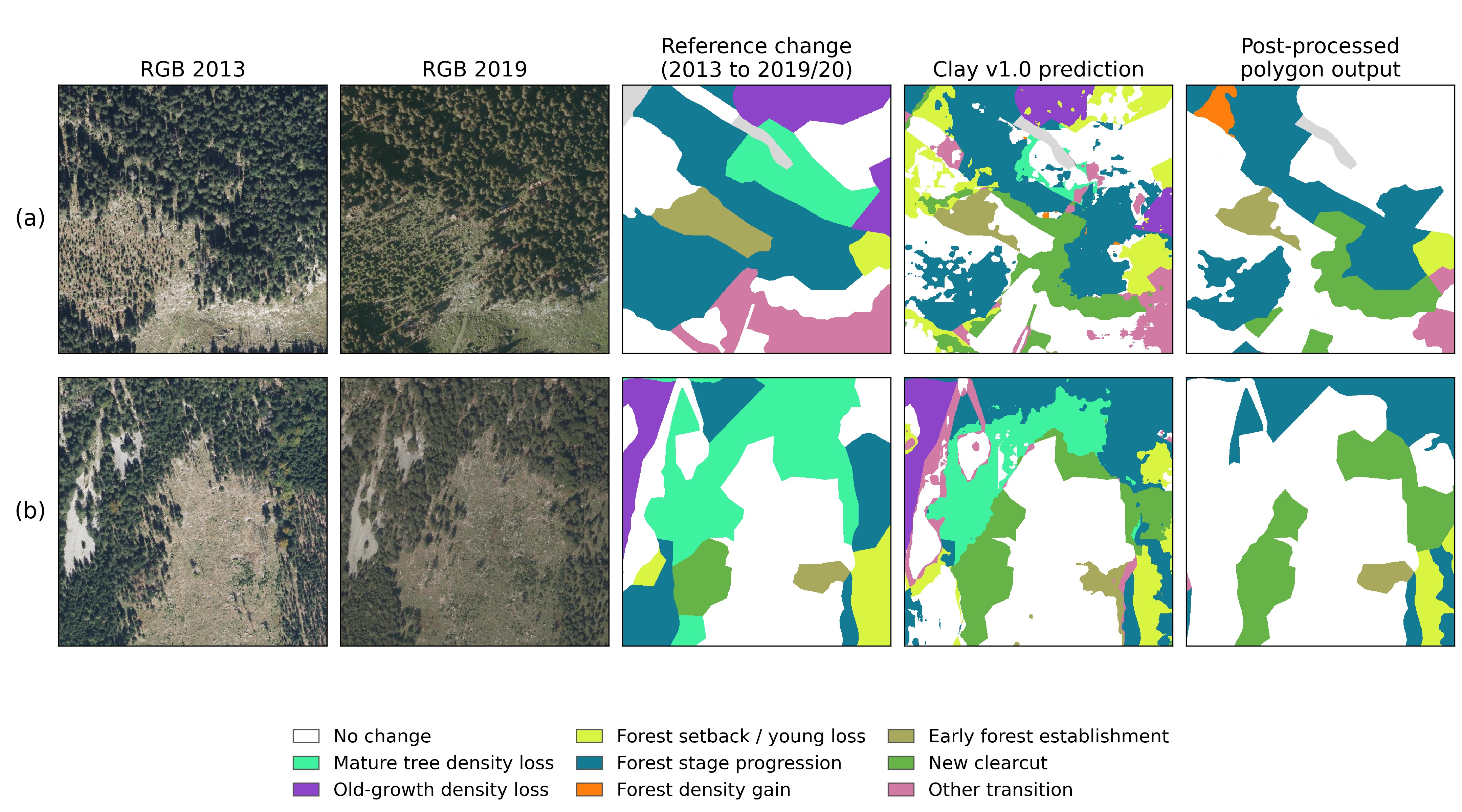}
    \caption{Cross-temporal (2013 to 2019/20) visual comparison for two 300\,m\,$\times$\,300\,m windows\linebreak (\textbf{a},\textbf{b}) inside the test patches: RGB orthophotos of 2013 and 2019 (columns 1--2), reference change map 2013 to 2019/20 (column 3), Clay v1.0 pixel prediction (column 4), and post-processed polygon output (column 5), both from the original Clay v1.0 run (original split), to which the operational post-processing was applied; grey marks unlabelled pixels (roads). Post-processing removes speckle and implausible transitions and snaps the prediction to the HabitAlp polygon geometry, but it can also remove correctly detected change, as for the mature-forest density loss in (\textbf{b}).\label{fig:visual_comparison}}
\end{figure}

\textls[-15]{The post-processing pipeline (Section~\ref{sec:pp_methods}; RQ4) is designed for operational usability. Physical-plausibility filtering reclassifies ecologically impossible transitions covering 9--15\% of the evaluated area, depending on the model (most commonly regression to a younger development stage) as ``no change'', yielding maps that are internally consistent with ecological succession. The full post-processing, applied to the original runs (one per model, original split), raises binary macro IoU for all six runs, whereas its effect on multi-class change is mixed (Table~\ref{tab:r1-b0_postproc}). Table~\ref{tab:r1-b5_postproc_steps} separates the steps for Clay v1.0 and the MiT-B2 U-Net. The six plausibility rules raise multi-class macro IoU by about 0.03 for both models, almost entirely through rule 3, which resets predicted regressions to a younger stage. Rule 3 is strict: it resets 74\,ha (Clay v1.0) and 152\,ha (MiT-B2 U-Net), of which only about 0.5\,ha are true stage regressions in the reference, so it raises the precision of forest setback at some cost in recall (Table~\ref{tab:r1-b5_rule3_cost}). The polygon step, vectorisation together with the canopy-cover update, then lowers multi-class macro IoU below that of the model output, while binary IoU stays above it, because the canopy-cover update moves predicted density loss into density gain or no change (Table~\ref{tab:r1-b5_class_recall}). The polygon step lowers recall mainly in the forest-loss classes: after it, recall of mature tree density loss, old-growth density loss, and forest setback is at most 0.17 for both models (0.19 to 0.53 in the model output), and over the whole workflow, binary change recall falls by 0.14 for Clay v1.0 and by 0.20 for the MiT-B2 U-Net, 0.08 of each in the polygon step (Table~\ref{tab:r1-b5_postproc_steps}). Figure~\ref{fig:visual_comparison} shows the resulting polygon product for two test windows.}

\begin{table}[H]
\caption{\textls[-25]{Effect of the full operational post-processing (plausibility rules, vectorisation, and canopy-cover update) on cross-temporal change-detection macro IoU (2013$\rightarrow$2019/20, RGB+NIR+nDSM, 0.2\,m). The post-processing was applied to the original runs (one per model, original split). $\Delta$ from unrounded values.}\label{tab:r1-b0_postproc}}
\begin{tabularx}{\textwidth}{lCCCCCC}
\toprule
& \multicolumn{3}{c}{\textbf{Binary}} & \multicolumn{3}{c}{\textbf{Multi-Class (9)}} \\
\cmidrule{2-7}
\textbf{Model} & \textbf{Before} & \textbf{After} & \boldmath{$\Delta$}  &  \textbf{Before} & \textbf{After} & \boldmath{$\Delta$}  \\
\midrule
Clay v1.0 & 0.431 & 0.448 & +0.017 &   0.240 & 0.207 & $-$0.033 \\
Clay v1.5 & 0.396 & 0.426 & +0.030 &   0.207 & 0.209 & +0.001 \\
Prithvi-EO-2.0 & 0.362 & 0.401 & +0.039 &  0.173 & 0.210 & +0.037 \\
DOFA & 0.406 & 0.439 & +0.033 &   0.207 & 0.220 & +0.013 \\
TerraMind & 0.369 & 0.389 & +0.019 &   0.196 & 0.187 & $-$0.010 \\
MiT-B2 U-Net & 0.420 & 0.424 & +0.004 &   0.212 & 0.211 & $-$0.001 \\
\bottomrule
\end{tabularx}
\end{table}

\begin{table}[H]

\caption{Cross-temporal change detection (2013$\rightarrow$2019/20, RGB+NIR+nDSM) after each post-processing stage: macro IoU over the nine change classes (Multi) and over change/no change (Bin.), binary change recall, and the area the stage resets to no change, at 0.2\,m over the evaluated 1159\,ha. Each rule is applied alone to the model output; the polygon step follows the six rules.\label{tab:r1-b5_postproc_steps}}
\begin{adjustwidth}{-\extralength}{0cm}
\begin{tabularx}{\fulllength}{lcccccccc}
\toprule
& \multicolumn{4}{c}{\textbf{Clay v1.0}} & \multicolumn{4}{c}{\textbf{MiT-B2 U-Net}} \\
\cmidrule{2-9}
\textbf{Stage} & \textbf{Reset (ha)} & \textbf{Multi IoU} & \textbf{Bin.\ IoU} & \textbf{Chg.\ rec.} & \textbf{Reset (ha)} & \textbf{Multi IoU} & \textbf{Bin.\ IoU} & \textbf{Chg.\ rec.} \\
\midrule
Model output (raw) & 0 & 0.240 & 0.431 & 0.83 &   0 & 0.212 & 0.420 & 0.73 \\
\midrule
Rule 1: non-forest $\rightarrow$ mature/old forest & 14.9 & 0.246 & 0.443 & 0.82 & 10.0 & 0.215 & 0.427 & 0.73 \\
Rule 2: young growth $\rightarrow$ old forest & 7.5 & 0.240 & 0.429 & 0.82 &   3.3 & 0.212 & 0.419 & 0.73 \\
Rule 3: regression to a younger stage & 74.1 & 0.267 & 0.462 & 0.78 & 152.0 & 0.240 & 0.459 & 0.62 \\
Rule 4: rock $\rightarrow$ grassland & 5.3 & 0.241 & 0.436 & 0.83 &  5.9 & 0.213 & 0.425 & 0.73 \\
Rule 5: water on slopes $>$35$^\circ$ & 3.4 & 0.240 & 0.431 & 0.82 & 4.6 & 0.212 & 0.420 & 0.73 \\
Rule 6: alpine vegetation below 650\,m & 1.1 & 0.240 & 0.432 & 0.82 &  0.5 & 0.212 & 0.420 & 0.73 \\
\midrule
All six rules & 106.2 & 0.275 & 0.478 & 0.77 &  176.3 & 0.245 & 0.469 & 0.61 \\
Polygon step (vectorisation + CC update) & - & 0.207 & 0.448 & 0.69 &  - & 0.211 & 0.424 & 0.53 \\
\bottomrule
\end{tabularx}
\end{adjustwidth}
\noindent{\footnotesize{The polygon step (vectorisation and LiDAR canopy-cover update) has its own no-data area (0.8--1.1\% fewer evaluated pixels); its reset area is not defined.}}

\end{table}
\vspace{-6pt}

\section{Discussion}
\label{sec:discussion}

\textbf{Detectability of habitat transitions (RQ1).} Major, abundant transitions (stage progression) and structurally distinct events (old-growth density loss) are detected with moderate accuracy, but rare and heterogeneous transitions remain poorly delineated by every model: some are over-predicted (high recall, low precision), others under-detected (low recall) by being absorbed into ``no change'' or confused with an adjacent development state. Class frequency is not the whole story: mature tree density loss covers 7.5\% of the test area but is mostly predicted as stage progression, so ecological ambiguity between adjacent development states limits detection even where labels are plentiful. New clearcut, although visually distinct, reaches only 0.18 IoU: Clay v1.0 finds half of the reference clearcut area, but it also labels open patches in grassland and in conifer forest below 80\% canopy cover as clearcut, which look like a fresh cut at 0.5\,m, so its precision stays at 0.21 (Table~\ref{tab:r1-b9_clay_perclass}). The inverse-area class weights add to the over-prediction of rare transitions, and a softer weighting (e.g.,\ square-root inverse frequency) would trade some recall for precision on these classes. The bottleneck is therefore label sparsity and this ecological ambiguity rather than the choice among the tested models. The achieved accuracies are a realistic baseline for this difficult task but not yet sufficient for fully automated operational use.

\textbf{GFMs versus the supervised baselines (RQ2).} Clay v1.0 is the only GFM ahead of the MiT-B2 U-Net, but its margin in multi-class change is small and only just robust (+0.024 macro IoU, lower CI bound +0.001), so a different test area or more seeds could remove it. In binary change and segmentation, the lead is clearer (lower CI bounds +0.013 and +0.006). The lead is also not uniform across classes, and Clay v1.0 needs about 45 times the computation of the MiT-B2 U-Net per patch. In practice, the difference between the two models is small, and a well-configured supervised U-Net remains a strong reference that no GFM surpasses by much. Against the convolutional ResNet-50 U-Net, in contrast, all five GFMs are robustly ahead in every metric, but so is the ImageNet-pretrained MiT-B2 U-Net, by a similar margin (+0.045 multi-class macro IoU). The gap to the CNN is therefore consistent with the transformer encoder rather than with Earth-observation pretraining, which adds at most a small gain, for Clay v1.0 only. Corley et al. \cite{Corley2024Change} found similar results: architectural sophistication often does not survive rigorous evaluation. GEO-Bench likewise reports that no single foundation model dominates across Earth-observation tasks~\cite{Lacoste2023GEOBench}, PANGAEA reaches the same conclusion for a broader set of tasks and finds supervised baselines competitive \cite{Marsocci2024PANGAEA}, and Nedungadi et al. \cite{Nedungadi2026Agriculture} observe unstable GFM rankings across agricultural datasets.

\textbf{Pretraining data composition.} The clearest contrast is between Clay v1.0 and the satellite-only Prithvi-EO-2.0, which is pretrained on 30\,m HLS data, a 60-fold resolution gap to the 0.5\,m grid of our models. In the post hoc comparison, Clay v1.0 is robustly ahead in all three metrics, and its pretraining corpus includes sub-metre aerial imagery (NAIP, LINZ). This is consistent with the hypothesis that the composition of pretraining data matters for very high-resolution ecological applications \cite{Roscher2024DataCentric}, but the comparison does not isolate it: the two models also differ in size (ViT-B vs. ViT-L), ViT patch size (8 vs. 16 pixels, i.e.,\ 4\,m vs. 8\,m tokens at 0.5\,m), pretraining objective, and band interface. Clay v1.0 versus v1.5 is the closest control for model size: v1.5 is a ViT-L trained on the same data family and has a lower mean in every metric, so a larger encoder alone does not help here. v1.5 also adds a DINOv2 representation loss, however, and our adaptation maps its bands through different platform metadata. DOFA also pretrains on NAIP aerial imagery, but in RGB only, and is level with the MiT-B2 U-Net, so including aerial data is not sufficient on its own; the proportion of aerial data, the band-adaptation mechanism, and the fine-tuning pipeline are likely to matter as well. For GFM developers, this points to aerial pretraining data as valuable for this domain, but only a controlled comparison, with the same architecture and objective trained with and without aerial data, can confirm it.

\textbf{Value of LiDAR (RQ3).} The nDSM gain is concentrated in the classes whose reference labels are themselves defined from LiDAR: the canopy-cover levels come from the same LiDAR surveys as the nDSM input, so the models can partly re-derive the label threshold. When the canopy-cover distinction is removed, the mean change-detection gain shrinks from +0.050 to +0.011 (Table~\ref{tab:r1-b3_cd_ablation}), and in 2019/20 segmentation the canopy-cover classes gain while the other classes lose (Table~\ref{tab:r1-b4_ndsm_ablation_mean}). The nDSM therefore mainly helps the models separate the two canopy-cover levels. After post-processing, whose canopy-cover update sets these levels per polygon from LiDAR, the nDSM input changes the mean scores by less than 0.01, so in the operational workflow LiDAR contributes mainly through the canopy-cover update. In 2019/20 the nDSM also lowers the scores of non-forest classes such as water, rock, and debris ($-0.093$ macro IoU, against $-0.012$ in 2013). One possible cause, which we did not test, is that the 2020 nDSM comes from a different LiDAR survey than the 2010/11 nDSM the models were trained on. None of the tested GFMs saw canopy height in pretraining (TerraMind uses elevation, but only as a 30\,m digital elevation model \cite{Jakubik2025TerraMind}), so every model has to learn the nDSM channel during fine-tuning.

\textbf{Operational usability (RQ4).} Rule-based post-processing turns the pixel predictions into maps that park staff can use directly: every remaining transition is ecologically possible, and the result follows the existing HabitAlp polygon geometry with LiDAR-based canopy cover. The plausibility rules were defined together with the national park staff on ecological grounds. Their effect on pixel metrics is small, with binary macro IoU changing by +0.00 to +0.04 and multi-class macro IoU by $-0.03$ to +0.04 across the six original runs (Table~\ref{tab:r1-b0_postproc}), but pixel metrics do not capture the main gain, a map that is consistent with forest succession and the park's own polygon data. The post-processing thus trades recall for precision: rule 3 alone raises forest-setback precision from 0.12 to 0.48 for Clay v1.0 (Table~\ref{tab:r1-b5_rule3_cost}). The price is lower recall for forest density loss and setback, which falls to at most 0.17 for both models, because majority voting on the 2013 polygons removes changes that cover only a small share of a polygon (Figure~\ref{fig:visual_comparison}b), rule 3 resets predicted stage regressions, and the canopy-cover update reassigns predicted density loss. The workflow also builds on the previous expert polygon map, which is always available when a map is updated, and on a current LiDAR survey, and steps (ii) and (iii) currently require proprietary software. Together with the limited accuracy on rare transitions (RQ1), this makes the post-processed map well suited for a semi-automated workflow, in which it flags candidate change polygons for expert verification and the model output, which retains more of the forest loss, can be consulted alongside it.

\textbf{Limitations.} The benchmark covers a single Alpine site, so the magnitude of differences may not transfer. It should be repeated on a contrasting eLTER site. The 2013 and 2019 orthophotos also differ in season, sensor, and radiometry; LiDAR coverage is only partial for 2020; and imagery and LiDAR of an epoch come from different years (2013 imagery with 2010/11 LiDAR, 2019 imagery with 2020 LiDAR; Table~\ref{tab:data_availability}). Part of the apparent accuracy drop between the two dates, and part of what is measured as change, therefore reflects these acquisition differences rather than genuine habitat change. Disentangling the two would require temporally denser, radiometrically harmonised imagery. Reference labels derive from expert interpretation and carry inherent uncertainty, especially at fuzzy class boundaries. Each model was trained with three seeds, the minimum that Corley et al. recommend for headline comparisons \cite{Corley2026SOTA}. The LiDAR ablation and the post-processing analysis, in contrast, rest on the single original runs (one per model, original split, earlier configuration, e.g.,\ batch size 8 for the MiT-B2 U-Net), whose CIs cover test-area sampling but not training randomness. Within each analysis, the two compared settings (with and without nDSM, or before and after post-processing) share the split and configuration, so these largely cancel in the reported effects. Training randomness does not cancel in this way: with seed standard deviations of up to 0.018 multi-class macro IoU (Table~\ref{tab:r1-b9_seed_summary}), single-run differences of a few hundredths can arise from it alone. We therefore draw conclusions from these analyses only where the effect is consistent across models (the nDSM gain in multi-class change and its concentration in the canopy-cover classes; the direction of each post-processing step for both models). A split control with the MiT-B2 U-Net scores lower on the original than on the masked split in multi-class change (0.183 vs. 0.209; Table~\ref{tab:r1-b9_seed_summary}), so the original split does not inflate these analyses, and the main caveats are the single runs and the earlier configuration. We do not interpret per-model values (reported for completeness) or the small residual nDSM gain with collapsed canopy-cover classes (+0.011), and the absolute scores of the original runs are not directly comparable with the benchmark tables. Learning rates were set per model from a learning-rate range test and a few preliminary runs, and each model kept one decoder, so the configurations are well tested but not exhaustively searched. No common height interface exists across the foundation models, so each receives the nDSM through its own band interface (Section~\ref{sec:models}), and part of the per-model differences in the LiDAR gain may come from this interface rather than from the height information itself. Because all models were trained on the full training set, the label-efficiency regime, where GFMs may show larger advantages, remains untested. The 2020 reference labels start from the 2013 polygon delineation: each polygon was re-interpreted for habitat change, and polygons were split into sub-polygons and given new classes where a change had occurred. Changes are therefore captured only down to the scale at which interpreters split polygons, whereas the models predict per pixel, so part of the pixel-level disagreement, including the moderate no-change recall (0.44), may reflect this difference in spatial granularity rather than model error alone. Two post-processing steps, vectorisation to the HabitAlp polygons and the per-polygon LiDAR canopy cover, were applied to the original runs alone and run in IMPACT Tools, proprietary software of JOANNEUM RESEARCH for in-house use only. We release their outputs, so the post-processing results (Table~\ref{tab:r1-b0_postproc}) can be recomputed, but the two steps themselves cannot be rerun from the repository.

\section{Conclusions}
\label{sec:conclusion}

We benchmarked five geospatial foundation models against two supervised U-Net baselines for fine-grained habitat change detection in an Alpine protected area, using very high-resolution imagery and LiDAR, an independent cross-temporal test set and three training seeds per model. The main findings are as follows: (1) Detectability (RQ1): Abundant and structurally distinct transitions (forest stage progression, old-growth density loss) are detected with moderate accuracy (IoU 0.38 to 0.40 for Clay v1.0), but no model reliably detects rare transitions, and forest setback and the residual ``other transition'' class stay below 0.10 IoU for every model. (2) GFMs versus the baselines (RQ2): The MiT-B2 U-Net and all five GFMs are robustly ahead of the convolutional ResNet-50 U-Net. Against the MiT-B2 U-Net, only Clay v1.0 is ahead (multi-class change macro IoU 0.233 vs. 0.209), by a small margin that is only just robust, and the other four GFMs are level with it. In a post hoc comparison, Clay v1.0 is also robustly ahead of the satellite-only Prithvi-EO-2.0 (0.233 vs. 0.204), which is consistent with its partly aerial pretraining data mattering; the two models, however, differ in size, patch size, and pretraining objective, and DOFA shows that aerial data alone are not sufficient. (3) Value of LiDAR (RQ3): The nDSM improves change detection for every model (+0.02 to +0.08 multi-class macro IoU), but the gain lies mostly in the canopy-cover classes, whose labels are derived from the same LiDAR, and shrinks to +0.011 on average when these distinctions are removed. (4) Operational usability (RQ4): Rule-based post-processing, defined together with the national park administration, converts predictions into ecologically plausible, polygon-structured maps compatible with park data standards, with a small effect on pixel accuracy (multi-class macro IoU $-0.03$ to +0.04) but lower recall for forest density loss and setback, so they are suited to flagging candidate changes for expert verification rather than to unsupervised use. When the full training set is available, a well-configured supervised U-Net therefore remains a strong reference for operational habitat monitoring in protected areas, and among the tested foundation models only the one pretrained partly on sub-metre aerial imagery adds a small gain, at a much higher computational cost. Repeating this benchmark on a contrasting eLTER site is the most pressing next step. Two further problems remain open: improving the detection of rare transitions, for example, through oversampling, synthetic augmentation, or class-aware losses, and pretraining GFMs on structurally informative data such as height and canopy structure.

\vspace{6pt}

\authorcontributions{Conceptualization, H.K. and M.H.; Methodology, H.K.; Software, H.K. and D.K.; Validation, H.K.; Formal Analysis, H.K.; Investigation, H.K.; Resources, M.H.; Data Curation, H.K.; Writing---Original Draft Preparation, H.K.; Writing---Review and Editing, M.H. and D.K.; Visualization, H.K.; Supervision, M.H.; Project Administration, M.H.; Funding Acquisition, M.H. All authors have read and agreed to the published version of the manuscript.}

\funding{This research was funded by the \"Osterreichische Forschungsf\"orderungsgesellschaft (FFG), grant number FO999911960, within the Austrian Space Application Program (ASAP). The authors acknowledge the financial support by the University of Graz.}%

\institutionalreview{Not applicable.}

\informedconsent{Not applicable.}

\dataavailability{The original data presented in the study, including the model outputs before and after post-processing, are openly available in HuggingFace (version v4.1) at %
 \url{https://huggingface.co/datasets/JR-DIGITAL/habitalp2.0} (accessed on 17 November 2026) \cite{Kristen2025HabitAlpData}. The code used in this study is openly available in GitHub at \url{https://github.com/JR-DIGITAL/habitalp_2} (accessed on 17 November 2026) \cite{Kristen2025HabitAlpCode}.}

\acknowledgments{The authors thank Nationalpark Ges\"ause GmbH for their collaboration and for providing the HabitAlp datasets and the INTERREG HabitAlp project for establishing the underlying land-cover nomenclature and long-term reference data dating back to 1954. During the preparation of this manuscript, the authors used Claude Opus 4.6 and, for the revision, Claude Opus 5.5 and Claude Fable 5.1 (Anthropic, San Francisco, CA, USA) for the purposes of drafting support, analysis scripting, and language editing. The authors have reviewed and edited the output and take full responsibility for the content of this publication.}

\conflictsofinterest{The authors declare no conflicts of interest. The funders had no role in the design of the study; in the collection, analyses, or interpretation of data; in the writing of the manuscript; or in the decision to publish the results.}

\abbreviations{Abbreviations}{The following abbreviations are used in this manuscript:\\

\noindent
\begin{tabular}{@{}ll}
GFM & Geospatial Foundation Model\\
CNN & Convolutional Neural Network\\
IoU & Intersection over Union\\
BA & Balanced Accuracy\\
nDSM & Normalised Digital Surface Model\\
RGB & Red, Green, Blue\\
NIR & Near-Infrared\\
CIR & Colour-Infrared\\
CD & Change Detection\\
CC & Canopy Cover\\
CI & Confidence Interval\\
GSD & Ground Sampling Distance\\
FCN & Fully Convolutional Network\\
ViT & Vision Transformer\\
HLS & Harmonized Landsat Sentinel-2
\end{tabular}}

\appendixtitles{yes}
\appendixstart
\appendix
\section{Class Scheme and Transition Rules}
\label{app:classes}

Table~\ref{tab:class_mapping} documents how the original HabitAlp habitat-type codes were consolidated into the 23 classes used for segmentation (Table~\ref{tab:classes}). Non-forest classes are assigned directly from the HabitAlp habitat-type code. Forest polygons (HabitAlp development-stage codes 7011--7030) are additionally split by dominant species group, using the coniferous and broad-leaved cover shares recorded in the HabitAlp species attributes (coniferous if the coniferous share is at least as large as the broad-leaved share), and, for the pole timber, mature, and old stages, by canopy cover (CC) derived from the 2010/11 LiDAR canopy-cover product (CC $\geq$ 80\% vs.\ CC $<$ 80\%). For the 2020 labels, canopy cover is derived in the same way from the 2020 LiDAR. Broad-leaved pole wood and both young-growth classes are not split by canopy cover because of their small extent. Mountain dwarf forest is assigned from the dwarf-pine (\emph{Pinus mugo}) cover share irrespective of the stage code. Clearcut polygons whose LiDAR canopy cover was still $\geq$40\%, i.e.,\ areas cut after the LiDAR acquisition, were masked as no-data to avoid contradictory height information.

\begin{table}[H]
\caption{Consolidation of HabitAlp habitat-type codes into the 23 segmentation classes. ID = class index used in Tables~\ref{tab:transition_rules} and \ref{tab:r1-b9_seg_perclass}; CC = LiDAR-derived canopy cover (\%).\label{tab:class_mapping}}
\tablesize{\small}
\begin{adjustwidth}{-\extralength}{0cm}
\begin{tabularx}{\fulllength}{rlm{5.cm}LL}
\toprule
\textbf{ID} & \textbf{Consolidated Class} & \textbf{HabitAlp 2013 Codes} & \textbf{Species Group} & \textbf{CC} \\
\midrule
1 & Waterbody & 2311, 2312, 2313, 2322, 2370, 2411, 2511, 2512, 2513, 2533 & - & - \\
2 & Gravel bank, shoal, fluviatile & 5410 & - & - \\
3 & Erosion area, gully & 5430, 5440 & - & - \\
4 & Debris-covered areas & 5700, 5701, 5702, 5703 & - & - \\
5 & Rock & 5800 & - & - \\
6 & Young coniferous (growth, thicket) & 7011, 7012 & coniferous & - \\
7 & Young broad-leaved (growth, thicket) & 7011, 7012 & broad-leaved & - \\
8 & Coniferous pole timber CC < 80 & 7013 & coniferous & $<$80 \\
9 & Coniferous pole timber CC $\geq$ 80 & 7013 & coniferous & $\geq$80 \\
10 & Broad-leaved pole wood & 7013 & broad-leaved & - \\
11 & Coniferous mature forest CC < 80 & 7014 & coniferous & $<$80 \\
12 & Coniferous mature forest CC $\geq$ 80 & 7014 & coniferous & $\geq$80 \\
13 & Broad-leaved mature forest CC < 80 & 7014 & broad-leaved & $<$80 \\
14 & Broad-leaved mature forest CC $\geq$ 80 & 7014 & broad-leaved & $\geq$80 \\
15 & Old coniferous forest, multilayered CC < 80 & 7015, 7021, 7022, 7030 & coniferous & $<$80 \\
16 & Old coniferous forest, multilayered CC $\geq$ 80 & 7015, 7021, 7022, 7030 & coniferous & $\geq$80 \\
17 & Old broad-leaved forest, multilayered CC < 80 & 7015, 7021, 7022, 7030 & broad-leaved & $<$80 \\
18 & Old broad-leaved forest, multilayered CC $\geq$ 80 & 7015, 7021, 7022, 7030 & broad-leaved & $\geq$80 \\
19 & Clearcut areas & 7710 (masked as no-data if CC $\geq$ 40) & - & - \\
20 & Mountain dwarf forest (`Krummholz' belt) & dwarf-pine cover share (species attribute), any stage code & - & - \\
21 & Grassland, buffer strip forest/open land & 4210, 4220, 4230, 4260, 4710, 4711, 4712, 4713, 4720, 4721, 4722, 4723, 4731, 4733, 7850, 9314 & - & - \\
22 & Alpine grassland, heath & 4240, 5610 & - & - \\
23 & Other minor classes & 67 remaining codes of low importance or small extent & - & - \\
\bottomrule
\end{tabularx}
\end{adjustwidth}
\end{table}

Change labels for 2013--2020 are derived from the two class maps by the rule set in Table~\ref{tab:transition_rules}, implemented in the function \texttt{get\_change\_array} of the public code repository~\cite{Kristen2025HabitAlpCode}. The rules operate on four forest development stages (young growth: IDs 6--7; pole wood: 8--10; mature forest: 11--14; old forest: 15--18), on the canopy-cover level (CC $\geq$ 80: IDs 9, 12, 14, 16, 18; CC $<$ 80: IDs 8, 11, 13, 15, 17), and on the non-forest vegetation classes (IDs 21--22). Mountain dwarf forest (ID 20) forms its own group: transitions between dwarf forest and the forest stages (IDs 6--18) are counted as ``other transition''. Rules are evaluated in the listed order; a pixel receives the label of the last rule that applies, and class changes matched by no specific rule remain in the residual ``other transition'' class. The same rules are applied to the reference maps and to the model predictions.

\begin{table}[H]
\caption{Rules aggregating class-to-class transitions between the 2013 and 2020 class maps into the nine change classes (class IDs as in Table~\ref{tab:class_mapping}).\label{tab:transition_rules}}

\begin{adjustwidth}{-\extralength}{0cm}
\begin{tabularx}{\fulllength}{lp{13.4cm}}
\toprule
\textbf{Change Class} & \textbf{Rule (2013 Class \boldmath{$\rightarrow$} 2020 Class)} \\
\midrule
No change & identical class at both dates \\
Other transition & default for every class change not matched by a rule below, e.g.,\ species-group changes within a stage, forest $\leftrightarrow$ non-forest conversions not covered by the establishment rule, mountain dwarf forest (20) $\leftrightarrow$ any forest stage (6--18), and changes between non-forest classes \\
New clearcut & any class $\rightarrow$ clearcut areas (19) \\
Early forest establishment & clearcut areas (19) $\rightarrow$ young growth (6, 7); non-forest vegetation (21, 22) $\rightarrow$ any forest stage (6--18) \\
Forest stage progression & young $\rightarrow$ pole; pole $\rightarrow$ mature; mature $\rightarrow$ old; pole $\rightarrow$ old \\
Forest setback/young loss & pole, mature, or old $\rightarrow$ young; old $\rightarrow$ mature; old $\rightarrow$ pole; mature $\rightarrow$ pole; CC $\geq$ 80 $\rightarrow$ CC $<$ 80 within pole wood (9 $\rightarrow$ 8) \\
Mature tree density loss & CC $\geq$ 80 $\rightarrow$ CC $<$ 80 within mature forest (12, 14 $\rightarrow$ 11, 13), irrespective of species group \\
Old-growth density loss & CC $\geq$ 80 $\rightarrow$ CC $<$ 80 within old forest (16, 18 $\rightarrow$ 15, 17), irrespective of species group \\
Forest density gain & CC $<$ 80 $\rightarrow$ CC $\geq$ 80 within the same stage (8 $\rightarrow$ 9; 11, 13 $\rightarrow$ 12, 14; 15, 17 $\rightarrow$ 16, 18) \\
\bottomrule
\end{tabularx}
\end{adjustwidth}
\end{table}

\section{Per-Class and Supplementary Results}
\label{app:perclass}

Tables~\ref{tab:r1-b9_seg_perclass} and \ref{tab:r1-b9_cd_perclass} complement the macro-averaged results in Section~\ref{sec:results} with per-class IoU for all seven models (revision runs, mean over three training seeds): single-date segmentation on the 2019/20 cross-temporal test patches and the 2013 to 2019/20 multi-class change detection. Table~\ref{tab:r1-b4_ndsm_ablation} gives the nDSM ablation per model, and Tables~\ref{tab:r1-b5_class_recall} and \ref{tab:r1-b5_rule3_cost} show the effect of the post-processing per change class and the cost of rule 3. Table~\ref{tab:r1-b9_seed_summary} lists all revision runs, including the decoder check and the split controls, and Table~\ref{tab:r1-b1_paired_tiles_only} the paired differences with the tile bootstrap alone. Table~\ref{tab:r1-b6_direct_cd} scores the direct change-detection maps of the conference version with the same evaluation; these models were trained on the only labelled change pair (2003 to 2013, RGB only) and are therefore not directly comparable with the post-classification models. %

\begin{table}[H]

\caption{Per-class IoU for single-date segmentation on the 2019/20 cross-temporal test patches: revision runs (masked split, best validation checkpoint), mean over three training seeds, RGB+NIR+nDSM, 0.2\,m. Best value per class in bold (ties at two decimals all bold).\label{tab:r1-b9_seg_perclass}}
\tablesize{\footnotesize}
\begin{adjustwidth}{-\extralength}{0cm}
\begin{tabularx}{\fulllength}{rlcccccCC}
\toprule
\textbf{ID} & \textbf{Class} & \textbf{Clay v1.0} & \textbf{Clay v1.5} & \textbf{Prithvi} & \textbf{DOFA} & \textbf{Terra\-Mind} & \begin{tabular}[b]{@{}c@{}}\textbf{MiT-B2}\\\textbf{U-Net}\end{tabular} & \begin{tabular}[b]{@{}c@{}}\textbf{ResNet-50}\\\textbf{U-Net}\end{tabular} \\
\midrule
1 & Waterbody & \textbf{0.25} & 0.24 & 0.19 & 0.11 & 0.22 & 0.24 & 0.16 \\
2 & Gravel bank, shoal, fluviatile & \textbf{0.15} & \textbf{0.15} & \textbf{0.15} & 0.07 & \textbf{0.15} & 0.13 & 0.05 \\
3 & Erosion area, gully & \textbf{0.33} & 0.26 & 0.23 & 0.22 & 0.24 & 0.30 & 0.26 \\
4 & Debris-covered areas & \textbf{0.33} & 0.29 & 0.28 & 0.23 & 0.22 & 0.28 & 0.21 \\
5 & Rock & \textbf{0.34} & 0.20 & 0.27 & 0.21 & 0.29 & 0.30 & 0.09 \\
6 & Young coniferous (growth, thicket) & \textbf{0.15} & 0.14 & \textbf{0.15} & 0.12 & 0.12 & \textbf{0.15} & 0.09 \\
7 & Young broad-leaved (growth, thicket) & 0.32 & 0.27 & \textbf{0.37} & 0.35 & 0.31 & 0.35 & 0.31 \\
8 & Coniferous pole timber CC < 80 & 0.15 & 0.14 & 0.14 & \textbf{0.20} & 0.16 & 0.13 & 0.15 \\
9 & Coniferous pole timber CC $\geq$ 80 & \textbf{0.28} & 0.23 & 0.24 & 0.26 & 0.25 & 0.23 & 0.20 \\
10 & Broad-leaved pole wood & \textbf{0.21} & 0.16 & 0.19 & 0.20 & 0.17 & 0.18 & 0.12 \\
11 & Coniferous mature forest CC < 80 & 0.21 & \textbf{0.25} & 0.23 & 0.16 & 0.24 & 0.17 & 0.14 \\
12 & Coniferous mature forest CC $\geq$ 80 & \textbf{0.25} & 0.22 & 0.22 & 0.24 & 0.23 & 0.16 & 0.21 \\
13 & Broad-leaved mature forest CC < 80 & 0.05 & 0.04 & 0.07 & 0.09 & 0.09 & 0.09 & \textbf{0.10} \\
14 & Broad-leaved mature forest CC $\geq$ 80 & 0.15 & 0.09 & 0.16 & \textbf{0.19} & \textbf{0.19} & 0.15 & 0.13 \\
15 & Old coniferous forest CC < 80 & \textbf{0.29} & 0.23 & 0.23 & 0.25 & 0.19 & 0.20 & 0.13 \\
16 & Old coniferous forest CC $\geq$ 80 & \textbf{0.17} & 0.09 & 0.11 & 0.16 & 0.08 & 0.12 & 0.11 \\
17 & Old broad-leaved forest CC < 80 & 0.08 & 0.08 & 0.08 & 0.06 & 0.06 & \textbf{0.09} & 0.04 \\
18 & Old broad-leaved forest CC $\geq$ 80 & 0.27 & 0.26 & 0.26 & 0.31 & \textbf{0.32} & 0.26 & 0.22 \\
19 & Clearcut areas & 0.28 & 0.27 & 0.25 & 0.24 & 0.21 & \textbf{0.30} & 0.20 \\
20 & Mountain dwarf forest & 0.45 & 0.37 & 0.38 & 0.36 & 0.37 & \textbf{0.47} & 0.13 \\
21 & Grassland, buffer strip & \textbf{0.34} & 0.27 & 0.31 & 0.26 & 0.27 & 0.30 & 0.26 \\
22 & Alpine grassland, heath & \textbf{0.27} & 0.18 & 0.18 & 0.12 & 0.15 & 0.24 & 0.11 \\
23 & Other minor classes & \textbf{0.43} & 0.37 & 0.29 & 0.23 & 0.26 & 0.35 & 0.21 \\
\midrule
 & Macro average & \textbf{0.25} & 0.21 & 0.22 & 0.20 & 0.21 & 0.23 & 0.16 \\
\bottomrule
\end{tabularx}
\end{adjustwidth}
\end{table}

\begin{table}[H]

\caption{Per-class IoU for cross-temporal multi-class change detection (2013$\rightarrow$2019/20, model output before post-processing): revision runs (masked split, best validation checkpoint), mean over three training seeds, RGB+NIR+nDSM, 0.2\,m. Best value per class in bold (ties at two decimals all~bold).\label{tab:r1-b9_cd_perclass}}
\tablesize{\small}
\begin{adjustwidth}{-\extralength}{0cm}
\begin{tabularx}{\fulllength}{lcccccCC}
\toprule
\textbf{Transition Class} & \textbf{Clay v1.0} & \textbf{Clay v1.5} & \textbf{Prithvi} & \textbf{DOFA} & \textbf{Terra\-Mind} & \begin{tabular}[b]{@{}c@{}}\textbf{MiT-B2}\\\textbf{U-Net}\end{tabular} & \begin{tabular}[b]{@{}c@{}}\textbf{ResNet-50}\\\textbf{U-Net}\end{tabular} \\
\midrule
No change & \textbf{0.39} & 0.33 & 0.35 & 0.34 & 0.35 & 0.35 & 0.28 \\
Forest stage progression & \textbf{0.40} & 0.38 & 0.36 & 0.37 & 0.34 & 0.35 & 0.28 \\
Old-growth density loss & \textbf{0.38} & 0.30 & 0.27 & 0.28 & 0.18 & 0.26 & 0.17 \\
New clearcut & 0.18 & 0.17 & 0.15 & 0.13 & 0.13 & \textbf{0.19} & 0.10 \\
Early forest establishment & 0.25 & 0.17 & 0.22 & 0.22 & \textbf{0.26} & 0.24 & 0.22 \\
Mature tree density loss & 0.19 & 0.26 & 0.24 & 0.20 & \textbf{0.27} & 0.20 & 0.17 \\
Forest density gain & 0.17 & 0.12 & 0.12 & \textbf{0.21} & 0.15 & 0.20 & 0.14 \\
Forest setback/young loss & 0.05 & 0.05 & 0.04 & \textbf{0.06} & 0.05 & 0.03 & 0.05 \\
Other transition & \textbf{0.09} & 0.08 & \textbf{0.09} & 0.07 & \textbf{0.09} & 0.07 & 0.06 \\
\midrule
Macro average & \textbf{0.23} & 0.21 & 0.20 & 0.21 & 0.20 & 0.21 & 0.16 \\
\bottomrule
\end{tabularx}
\end{adjustwidth}
\end{table}

\vspace{-6pt}
\begin{table}[H]

\caption{Effect of the LiDAR nDSM on single-date segmentation by class group: macro IoU of RGB+NIR+nDSM (E) and RGB+NIR (F) and the difference $\Delta$ = E $-$ F, at 0.2\,m. 2019/20: cross-temporal test patches; 2013: in-domain test blocks. 95\% CI of $\Delta$: paired spatial block bootstrap over 500\,m tiles, 1000 resamples (2019/20: 88 tiles; 2013: 178 tiles); the model mean uses the same resamples for all models. The CIs cover test-set sampling only, not training randomness (single runs).\label{tab:r1-b4_ndsm_ablation}}
\tablesize{\footnotesize}
\begin{adjustwidth}{-\extralength}{0cm}
\begin{tabularx}{\fulllength}{lCCCcCCCc}
\toprule
& \multicolumn{4}{c}{\textbf{2019/20 (Cross-Temporal)}} & \multicolumn{4}{c}{\textbf{2013 (in-Domain)}} \\
\cmidrule{2-9}
\textbf{Model} & \textbf{E} & \textbf{F} & \boldmath{$\Delta$} & \textbf{[95\% CI]}  & \textbf{E} & \textbf{F} & \boldmath{$\Delta$} & \textbf{[95\% CI]} \\
\midrule
\multicolumn{9}{l}{\textit{All 23 classes}} \\
Clay v1.0 & 0.252 & 0.269 & $-$0.017 & [$-$0.033, $-$0.001]  & 0.327 & 0.320 & +0.006 & [$-$0.003, +0.016] \\
Clay v1.5 & 0.219 & 0.233 & $-$0.014 & [$-$0.032, +0.002]   & 0.305 & 0.288 & +0.017 & [+0.008, +0.028] \\
Prithvi-EO-2.0 & 0.166 & 0.225 & $-$0.059 & [$-$0.076, $-$0.040]   & 0.263 & 0.299 & $-$0.036 & [$-$0.048, $-$0.023] \\
DOFA & 0.219 & 0.177 & +0.042 & [+0.020, +0.060]  & 0.301 & 0.272 & +0.030 & [+0.019, +0.040] \\
TerraMind & 0.200 & 0.205 & $-$0.005 & [$-$0.027, +0.015]   & 0.279 & 0.290 & $-$0.011 & [$-$0.020, 0.000] \\
MiT-B2 U-Net & 0.265 & 0.266 & $-$0.001 & [$-$0.017, +0.012]   & 0.322 & 0.309 & +0.013 & [+0.005, +0.021] \\
\midrule
Mean (6 models) & 0.220 & 0.229 & $-$0.009 & [$-$0.024, +0.004] & 0.300 & 0.296 & +0.003 & [$-$0.002, +0.009] \\
\midrule
\multicolumn{9}{l}{\textit{10 canopy-cover classes (CC $<$ 80 / CC $\geq$ 80)}} \\
Clay v1.0 & 0.201 & 0.161 & +0.041 & [+0.016, +0.064]   & 0.218 & 0.210 & +0.007 & [$-$0.009, +0.024] \\
Clay v1.5 & 0.173 & 0.141 & +0.031 & [+0.014, +0.049]  & 0.207 & 0.192 & +0.016 & [+0.001, +0.031] \\
Prithvi-EO-2.0 & 0.150 & 0.125 & +0.025 & [+0.003, +0.050]  & 0.174 & 0.193 & $-$0.019 & [$-$0.033, $-$0.004] \\
DOFA & 0.210 & 0.105 & +0.105 & [+0.079, +0.129]   & 0.202 & 0.168 & +0.034 & [+0.017, +0.049] \\
TerraMind & 0.193 & 0.111 & +0.082 & [+0.062, +0.101]   & 0.179 & 0.182 & $-$0.003 & [$-$0.020, +0.015] \\
MiT-B2 U-Net & 0.189 & 0.160 & +0.029 & [+0.010, +0.046]  & 0.212 & 0.200 & +0.011 & [$-$0.002, +0.023] \\
\midrule
Mean (6 models) & 0.186 & 0.134 & +0.052 & [+0.037, +0.067]  & 0.199 & 0.191 & +0.008 & [0.000, +0.016] \\
\midrule
\multicolumn{9}{l}{\textit{Other 13 classes}} \\
Clay v1.0 & 0.291 & 0.352 & $-$0.062 & [$-$0.082, $-$0.042] & 0.411 & 0.405 & +0.006 & [$-$0.006, +0.017] \\
Clay v1.5 & 0.255 & 0.304 & $-$0.049 & [$-$0.077, $-$0.024]   & 0.381 & 0.362 & +0.019 & [+0.007, +0.030] \\
Prithvi-EO-2.0 & 0.179 & 0.302 & $-$0.123 & [$-$0.145, $-$0.100]   & 0.332 & 0.381 & $-$0.049 & [$-$0.067, $-$0.033] \\
DOFA & 0.226 & 0.233 & $-$0.007 & [$-$0.036, +0.017]   & 0.378 & 0.351 & +0.026 & [+0.014, +0.039] \\
TerraMind & 0.206 & 0.277 & $-$0.071 & [$-$0.101, $-$0.046]  & 0.356 & 0.372 & $-$0.016 & [$-$0.030, $-$0.002] \\
MiT-B2 U-Net & 0.323 & 0.347 & $-$0.024 & [$-$0.044, $-$0.006]  & 0.408 & 0.393 & +0.015 & [+0.006, +0.024] \\
\midrule
Mean (6 models) & 0.246 & 0.303 & $-$0.056 & [$-$0.075, $-$0.040]   & 0.377 & 0.377 & 0.000 & [$-$0.007, +0.007] \\
\midrule
\multicolumn{9}{l}{\textit{Canopy-cover pairs merged (5 stage classes)}} \\
Clay v1.0 & 0.275 & 0.276 & $-$0.001 & [$-$0.038, +0.037]   & 0.288 & 0.300 & $-$0.012 & [$-$0.037, +0.016] \\
Clay v1.5 & 0.250 & 0.272 & $-$0.022 & [$-$0.048, +0.004]   & 0.286 & 0.267 & +0.019 & [+0.003, +0.035] \\
Prithvi-EO-2.0 & 0.246 & 0.246 & +0.001 & [$-$0.033, +0.038]   & 0.256 & 0.280 & $-$0.025 & [$-$0.048, +0.001] \\
DOFA & 0.289 & 0.203 & +0.086 & [+0.052, +0.121]  & 0.276 & 0.240 & +0.036 & [+0.017, +0.054] \\
TerraMind & 0.270 & 0.178 & +0.092 & [+0.060, +0.123]   & 0.256 & 0.268 & $-$0.012 & [$-$0.037, +0.016] \\
MiT-B2 U-Net & 0.265 & 0.318 & $-$0.053 & [$-$0.080, $-$0.026]   & 0.300 & 0.297 & +0.002 & [$-$0.013, +0.018] \\
\midrule
Mean (6 models) & 0.266 & 0.249 & +0.017 & [$-$0.005, +0.040]   & 0.277 & 0.275 & +0.001 & [$-$0.012, +0.015] \\
\bottomrule
\end{tabularx}
\end{adjustwidth}
\noindent{\footnotesize{Canopy-cover classes: 8, 9, 11--18 (pole, mature and old forest split at 80\% LiDAR canopy cover). Other: classes 1--7, 10, 19--23. Merged: each CC$<$80 / CC$\geq$80 pair counted as one stage class.}}

\end{table}

\begin{table}[H]
\caption{Recall per change class (omission = 1 $-$ recall) for the model output (Raw), after the six plausibility rules (Rules) and after the polygon step (Polygon); 2013$\rightarrow$2019/20, RGB+NIR+nDSM, 0.2\,m.\label{tab:r1-b5_class_recall}}
\tablesize{\small}
\begin{tabularx}{\textwidth}{lCCCCCC}
\toprule
& \multicolumn{3}{c}{\textbf{Clay v1.0}} &  \multicolumn{3}{c}{\textbf{MiT-B2 U-Net}} \\
\cmidrule{2-7}
\textbf{Change Class} & \textbf{Raw} & \textbf{Rules} & \textbf{Polygon} & \textbf{Raw} & \textbf{Rules} & \textbf{Polygon} \\
\midrule
No change & 0.45 & 0.56 & 0.57 & 0.49 & 0.67 & 0.66 \\
Mature tree density loss & 0.19 & 0.19 & 0.06 & 0.28 & 0.28 & 0.17 \\
Old-growth density loss & 0.53 & 0.53 & 0.11 & 0.38 & 0.38 & 0.17 \\
Forest setback/young loss & 0.49 & 0.43 & 0.15 & 0.38 & 0.30 & 0.15 \\
Forest stage progression & 0.61 & 0.61 & 0.65 & 0.38 & 0.38 & 0.39 \\
Forest density gain & 0.31 & 0.31 & 0.77 & 0.21 & 0.21 & 0.87 \\
Early forest establishment & 0.65 & 0.65 & 0.70 & 0.63 & 0.63 & 0.65 \\
New clearcut & 0.59 & 0.59 & 0.65 & 0.46 & 0.46 & 0.48 \\
Other transition & 0.42 & 0.38 & 0.34 & 0.49 & 0.46 & 0.44 \\
\bottomrule
\end{tabularx}
\end{table}
\vspace{-6pt}
\begin{table}[H]

\caption{Cost of plausibility rule 3 (regression to a younger forest stage) at 0.2\,m, 2013$\rightarrow$2019/20, RGB+NIR+nDSM: the area it resets to no change, split by the reference class, and forest-setback precision, recall, and IoU before and after the rule. Evaluated area 1159\,ha; reference forest setback 22.5\,ha, of which 0.72\,ha are true stage regressions (forest stage 2013 $>$ 2019/20).\label{tab:r1-b5_rule3_cost}}

\begin{tabularx}{\textwidth}{lCC}
\toprule
& \textbf{Clay v1.0} & \textbf{MiT-B2 U-Net} \\
\midrule
Area reset by rule 3 (ha) & 74.1 & 152.0 \\
\quad share of the evaluated area (\%) & 6.4 & 13.1 \\
\quad no change in the reference (ha) & 53.2 & 99.9 \\
\quad other real change in the reference (ha) & 20.4 & 51.7 \\
\quad true stage regression in the reference (ha) & 0.54 & 0.49 \\
\midrule
Setback precision, model output & 0.12 & 0.05 \\
Setback precision, after rule 3 & 0.48 & 0.43 \\
Setback recall, model output & 0.49 & 0.38 \\
Setback recall, after rule 3 & 0.43 & 0.30 \\
Setback IoU, model output & 0.10 & 0.05 \\
Setback IoU, after rule 3 & 0.29 & 0.21 \\
\bottomrule
\end{tabularx}

\end{table}
\vspace{-6pt}
\begin{table}[H]

\caption{Revision runs on the masked split: macro IoU at 0.2\,m, mean $\pm$ standard deviation over training seeds (best validation checkpoint). Seg.: single-date segmentation (23 classes) on the 2019/20 cross-temporal test patches and the 2013 in-domain test blocks; CD: 2013$\rightarrow$2019/20 change detection of the model output (before post-processing), binary and nine classes.\label{tab:r1-b9_seed_summary}}
\begin{adjustwidth}{-\extralength}{0cm}
\begin{tabularx}{\fulllength}{lcCCCC}
\toprule
\textbf{Model} & \textbf{Seeds} & \textbf{Seg.\ 2019/20} & \textbf{Seg.\ 2013} & \textbf{CD Binary} & \textbf{CD Multi-Class} \\
\midrule
Clay v1.0 & 3 & 0.249 $\pm$ 0.003 & 0.329 $\pm$ 0.003 & 0.425 $\pm$ 0.007 & 0.233 $\pm$ 0.003 \\
Clay v1.5 & 3 & 0.210 $\pm$ 0.006 & 0.311 $\pm$ 0.002 & 0.386 $\pm$ 0.011 & 0.209 $\pm$ 0.005 \\
Prithvi-EO-2.0 & 3 & 0.216 $\pm$ 0.013 & 0.304 $\pm$ 0.013 & 0.394 $\pm$ 0.012 & 0.204 $\pm$ 0.009 \\
DOFA & 3 & 0.201 $\pm$ 0.004 & 0.294 $\pm$ 0.003 & 0.395 $\pm$ 0.009 & 0.210 $\pm$ 0.013 \\
TerraMind & 3 & 0.208 $\pm$ 0.003 & 0.306 $\pm$ 0.005 & 0.388 $\pm$ 0.007 & 0.204 $\pm$ 0.005 \\
MiT-B2 U-Net & 3 & 0.226 $\pm$ 0.015 & 0.312 $\pm$ 0.009 & 0.389 $\pm$ 0.020 & 0.209 $\pm$ 0.018 \\
ResNet-50 U-Net & 3 & 0.157 $\pm$ 0.014 & 0.285 $\pm$ 0.010 & 0.344 $\pm$ 0.017 & 0.164 $\pm$ 0.014 \\
\midrule
\multicolumn{6}{l}{\textit{Checks and controls}} \\
Clay v1.0, U-Net decoder & 1 & 0.224 & 0.321 & 0.408 & 0.227 \\
Prithvi-EO-2.0, U-Net decoder & 1 & 0.202 & 0.313 & 0.396 & 0.221 \\
TerraMind, LR $5\times10^{-5}$ & 1 & 0.219 & 0.308 & 0.400 & 0.214 \\
MiT-B2 U-Net, original split & 3 & 0.228 $\pm$ 0.002 & 0.313 $\pm$ 0.005 & 0.385 $\pm$ 0.010 & 0.183 $\pm$ 0.012 \\
MiT-B2 U-Net, block-drop split & 3 & 0.215 $\pm$ 0.006 & 0.312 $\pm$ 0.001 & 0.378 $\pm$ 0.006 & 0.203 $\pm$ 0.012 \\
\bottomrule
\end{tabularx}
\end{adjustwidth}
\end{table}

\begin{table}[H]

\caption{Paired differences as in Table~\ref{tab:r1-b1_paired}, with unadjusted 95\% CI from a paired bootstrap over the 88 test tiles (500\,m) only (10,000 replicates): these intervals keep the trained seeds fixed and cover test-tile sampling only. They are about half as wide (0.4 to 0.8 times the width) as the main intervals (tiles and seeds) in Table~\ref{tab:r1-b1_paired}. Bold: the 95\% CI excludes zero. $^\dagger$ Post hoc.\label{tab:r1-b1_paired_tiles_only}}
\tablesize{\footnotesize}
\begin{adjustwidth}{-\extralength}{0cm}
\begin{tabularx}{\fulllength}{Lrlrlrl}
\toprule
Comparison & \multicolumn{2}{c}{\textbf{Seg.\ 2019/20}} & \multicolumn{2}{c}{\textbf{CD Binary}} & \multicolumn{2}{c}{\textbf{CD Multi-Class}} \\
\midrule
Clay v1.0 $-$ MiT-B2 U-Net & \textbf{+0.023} & \textbf{[+0.014, +0.032]} & \textbf{+0.036} & \textbf{[+0.025, +0.048]} & \textbf{+0.024} & \textbf{[+0.009, +0.037]} \\
Clay v1.5 $-$ MiT-B2 U-Net & \textbf{$-$0.017} & \textbf{[$-$0.026, $-$0.007]} & $-$0.003 & [$-$0.017, +0.010] & $-$0.001 & [$-$0.012, +0.010] \\
Prithvi-EO-2.0 $-$ MiT-B2 U-Net & \textbf{$-$0.010} & \textbf{[$-$0.021, $-$0.000]} & +0.006 & [$-$0.006, +0.017] & $-$0.005 & [$-$0.018, +0.009] \\
DOFA $-$ MiT-B2 U-Net & \textbf{$-$0.025} & \textbf{[$-$0.036, $-$0.014]} & +0.006 & [$-$0.008, +0.019] & 0.000 & [$-$0.018, +0.019] \\
TerraMind $-$ MiT-B2 U-Net & \textbf{$-$0.018} & \textbf{[$-$0.030, $-$0.007]} & 0.000 & [$-$0.016, +0.014] & $-$0.006 & [$-$0.020, +0.011] \\
ResNet-50 U-Net $-$ MiT-B2 U-Net & \textbf{$-$0.069} & \textbf{[$-$0.082, $-$0.054]} & \textbf{$-$0.045} & \textbf{[$-$0.063, $-$0.028]} & \textbf{$-$0.045} & \textbf{[$-$0.058, $-$0.031]} \\
Clay v1.0 $-$ Prithvi-EO-2.0 $^\dagger$ & \textbf{+0.033} & \textbf{[+0.025, +0.042]} & \textbf{+0.031} & \textbf{[+0.018, +0.043]} & \textbf{+0.029} & \textbf{[+0.014, +0.041]} \\
\midrule
\multicolumn{7}{l}{\textit{Against the ResNet-50 U-Net}} \\
Clay v1.0 $-$ ResNet-50 U-Net $^\dagger$ & \textbf{+0.092} & \textbf{[+0.076, +0.106]} & \textbf{+0.081} & \textbf{[+0.063, +0.100]} & \textbf{+0.069} & \textbf{[+0.053, +0.082]} \\
Clay v1.5 $-$ ResNet-50 U-Net $^\dagger$ & \textbf{+0.052} & \textbf{[+0.040, +0.063]} & \textbf{+0.042} & \textbf{[+0.027, +0.057]} & \textbf{+0.044} & \textbf{[+0.031, +0.057]} \\
Prithvi-EO-2.0 $-$ ResNet-50 U-Net $^\dagger$ & \textbf{+0.059} & \textbf{[+0.046, +0.068]} & \textbf{+0.050} & \textbf{[+0.037, +0.064]} & \textbf{+0.040} & \textbf{[+0.027, +0.054]} \\
DOFA $-$ ResNet-50 U-Net $^\dagger$ & \textbf{+0.044} & \textbf{[+0.031, +0.055]} & \textbf{+0.051} & \textbf{[+0.039, +0.062]} & \textbf{+0.046} & \textbf{[+0.031, +0.059]} \\
TerraMind $-$ ResNet-50 U-Net $^\dagger$ & \textbf{+0.051} & \textbf{[+0.041, +0.059]} & \textbf{+0.044} & \textbf{[+0.029, +0.059]} & \textbf{+0.040} & \textbf{[+0.026, +0.056]} \\
\midrule
\multicolumn{7}{l}{\textit{Supporting comparisons}} \\
Clay v1.0, U-Net decoder $-$ Prithvi-EO-2.0, U-Net decoder & \textbf{+0.022} & \textbf{[+0.009, +0.035]} & +0.012 & [$-$0.007, +0.031] & +0.006 & [$-$0.009, +0.024] \\
MiT-B2 U-Net $-$ MiT-B2 U-Net, original split & $-$0.001 & [$-$0.010, +0.006] & +0.004 & [$-$0.008, +0.015] & \textbf{+0.026} & \textbf{[+0.012, +0.040]} \\
MiT-B2 U-Net $-$ MiT-B2 U-Net, block-drop split & \textbf{+0.012} & \textbf{[+0.005, +0.018]} & \textbf{+0.011} & \textbf{[+0.002, +0.020]} & +0.006 & [$-$0.003, +0.018] \\
MiT-B2 U-Net, original split $-$ MiT-B2 U-Net, block-drop split & \textbf{+0.013} & \textbf{[+0.003, +0.023]} & +0.007 & [$-$0.005, +0.019] & \textbf{$-$0.020} & \textbf{[$-$0.032, $-$0.006]} \\
\bottomrule
\end{tabularx}
\end{adjustwidth}
\end{table}
\vspace{-6pt}
\begin{table}[H]

\caption{Direct change-detection maps scored with the evaluation of this study (2013$\rightarrow$2019/20, reference change map at 0.2\,m, same test area and metric code), next to the post-classification models. Macro IoU, macro F1 and balanced accuracy (BA); 95\% CI of IoU: spatial block bootstrap over 500\,m tiles, 1000 resamples (88 tiles).\label{tab:r1-b6_direct_cd}}
\begin{adjustwidth}{-\extralength}{0cm}
\begin{tabularx}{\fulllength}{lll CcCC CcCC}
\toprule
& &  &\multicolumn{4}{c}{\textbf{Binary}} & \multicolumn{4}{c}{\textbf{Multi-Class (9)}} \\
\cmidrule{4-11}
\textbf{Method} & \textbf{Output} & \textbf{Input} & \textbf{IoU} & \textbf{[95\% CI]} & \textbf{F1} & \textbf{BA} &  \textbf{IoU} & \textbf{[95\% CI]} & \textbf{F1} & \textbf{BA} \\
\midrule
\multicolumn{11}{l}{\textit{Direct change detection (2025 models)}} \\
ChangeViT & binary & RGB & 0.319 & [0.286, 0.350] & 0.434 & 0.499 &  - & - & - & - \\
U-Net, MiT-B2 encoder & binary & RGB & 0.326 & [0.292, 0.359] & 0.441 & 0.510   & - & - & - & - \\
U-Net, EfficientNet-B4 encoder & 9 classes \textsuperscript{a} & RGB & 0.346 & [0.326, 0.367] & 0.478 & 0.520 & 0.104 & [0.090, 0.118] & 0.157 & 0.198 \\
\midrule
\multicolumn{11}{l}{\textit{Post-classification (this study, model output before post-processing)}} \\
Clay v1.0 & 9 classes & RGB+NIR+nDSM & 0.431 & [0.405, 0.457] & 0.602 & 0.637 &   0.240 & [0.209, 0.266] & 0.370 & 0.473 \\
Clay v1.0 & 9 classes & RGB+NIR & 0.375 & [0.352, 0.399] & 0.546 & 0.557 &   0.177 & [0.152, 0.196] & 0.289 & 0.406 \\
MiT-B2 U-Net & 9 classes & RGB+NIR+nDSM & 0.420 & [0.396, 0.446] & 0.592 & 0.613  & 0.212 & [0.175, 0.241] & 0.334 & 0.411 \\
MiT-B2 U-Net & 9 classes & RGB+NIR & 0.392 & [0.372, 0.415] & 0.563 & 0.578 & 0.168 & [0.146, 0.184] & 0.269 & 0.372 \\
\bottomrule
\end{tabularx}
\end{adjustwidth}
\noindent{\footnotesize{\textsuperscript{a} The 9-class map collapsed to change/no change for the binary columns. Direct models were trained on RGB image pairs (2003 has no CIR) in 2025. AGIT~v2 reported for them macro IoU ChangeViT binary 0.36; U-Net, EfficientNet-B4 encoder binary 0.38; U-Net, EfficientNet-B4 encoder multi-class 0.10. These values differ because this table uses the revised v4 reference change map and test area of all other rows, whereas the AGIT~v2 direct rows reproduce only against an earlier reference with a smaller share of changed pixels; its binary U-Net row also refers to the nine-class model collapsed to change/no change.}}

\end{table}

\section{Model Selection}
\label{app:model-selection}

The GFM set was fixed in March 2026 from the pretrained models in TerraTorch~1.2.4 and TorchGeo~0.9.0, the versions of the original runs (Table~\ref{tab:model-selection}). TerraTorch~1.2.13, which we used for the revision runs and which is the latest release (August 2026), provides the same segmentation backbones. Corley et al. \cite{Corley2026SOTA} argue that a comparison does not need to include every available backbone. We required open weights and a TerraTorch segmentation factory so that all models share one training pipeline. All five selected GFMs saw near-infrared bands during pretraining, Clay also in aerial NAIP imagery, whereas the very high-resolution weights among the alternatives (SatlasPretrain NAIP and DINOv3 satellite) are RGB-only. SatlasPretrain trains separate weights per sensor, and its multispectral Sentinel-2 weights, like SSL4EO-S12 and SSL4EO-L, come from 10 to 30\,m satellite imagery only. TorchGeo additionally provides Scale-MAE \cite{Reed2023ScaleMAE}, CROMA~\cite{Fuller2023CROMA}, Copernicus-FM \cite{Wang2025CopernicusFM}, Panopticon \cite{Waldmann2025Panopticon}, and the pixel time-series models Presto \cite{Tseng2023Presto} and TESSERA \cite{Feng2025TESSERA}, which have no TerraTorch segmentation factory. We limited the set to five GFMs because each needs model-specific band adaptation and learning-rate exploration.

\begin{table}[H]

\caption{GFM backbones with a segmentation factory in TerraTorch~1.2.4 and their pretrained weights.\label{tab:model-selection}}
\begin{adjustwidth}{-\extralength}{0cm}
\scriptsize
\setlength{\tabcolsep}{4pt}
\renewcommand{\arraystretch}{1.15}
\begin{tabularx}{\linewidth}{>{\raggedright\arraybackslash}p{3.3cm}>{\raggedright\arraybackslash}X>{\raggedright\arraybackslash}p{4.2cm}c}
\toprule
\textbf{Model} & \textbf{Pretraining Data (GSD)} & \textbf{Input Bands of the Weights} & \textbf{Used} \\
\midrule
Clay v1.0/v1.5 \cite{Clay2023Clay} & Sentinel-1/2, Landsat 8/9 (10--30\,m), NAIP (1\,m), LINZ (0.5\,m); v1.5 also MODIS & any; NAIP: R, G, B, NIR & yes \\
Prithvi-EO-2.0 \cite{Szwarcman2025Prithvi} & HLS (30\,m) & B, G, R, NIR, SWIR1, SWIR2 & yes \\
DOFA \cite{Xiong2024DOFA} & Sentinel-1/2 (10\,m), Gaofen-2 (4\,m), NAIP (1\,m, RGB), EnMAP (30\,m) & any (wavelength-conditioned) & yes \\
TerraMind \cite{Jakubik2025TerraMind} & Sentinel-1/2, Copernicus DEM (on a 10\,m grid) & S2: 12 bands; S1; DEM & yes \\
SatlasPretrain Swin-B \cite{Bastani2023SatlasPretrain} & separate weights: NAIP (1\,m) or Sentinel-2 (10\,m) & NAIP: R, G, B; S2: 9 bands incl.\ NIR & no \\
DINOv3 (SAT-493M) \cite{Simeoni2025DINOv3} & Maxar (0.6\,m) & R, G, B & no \\
SSL4EO-S12 \cite{Wang2023SSL4EOS12} & Sentinel-2 (10\,m) & 13 bands & no \\
SSL4EO-L \cite{Stewart2023SSL4EOL} & Landsat 4--9 (30\,m) & 6--11 bands & no \\
\bottomrule
\end{tabularx}
\end{adjustwidth}
\end{table}

\section{Training Details}
\label{app:training}

\textls[-25]{Table~\ref{tab:r1-b8_training_facts} lists the training facts of every revision run and of the original runs, Table~\ref{tab:r1-b8_lr_table} the learning rates explored per model, Table~\ref{tab:r1-b9_best_vs_last} the test scores of the best validation checkpoint next to those of the last epoch, and Table~\ref{tab:r1-b10_efficiency} the model size and computational cost.} %

\begin{table}[H]

\caption{Training facts per run. Revision runs: batch size 4, 512 training batches per epoch, 256~validation patches per epoch, a fixed budget of 200 epochs without early stopping (10-epoch linear warm-up, then cosine decay to $10^{-7}$), and the checkpoint with the best validation macro IoU; values per seed separated by slashes. Runs of the submitted manuscript: up to 150 epochs with early stopping (patience 40), 64 validation batches per epoch.\label{tab:r1-b8_training_facts}}
\scriptsize
\begin{adjustwidth}{-\extralength}{0cm}
\setlength{\tabcolsep}{6.4pt}
\centering
\begin{threeparttable}
\begin{tabular}{llcccccccl}
\toprule
\textbf{Model} & \textbf{Split} & \textbf{Seeds} & \textbf{Batch} & \textbf{Patches/ep.} & \textbf{Epochs} & \textbf{Best Epoch} & \textbf{Best val IoU} & \textbf{Last val IoU} & \textbf{Stop} \\
\midrule
\multicolumn{10}{l}{\textit{Revision runs (masked split unless stated)}} \\
Clay v1.0 (FCN) & masked & 0,1,2 & 4 & 2048 & 200 & 162/194/160 & 0.311/0.314/0.308 & 0.307/0.254/0.246 & fixed budget \\
\quad U-Net decoder & masked & 0 & 4 & 2048 & 200 & 129 & 0.307 & 0.287 & fixed budget \\
Clay v1.5 (FCN) & masked & 0,1,2 & 4 & 2048 & 200 & 199/162/118 & 0.304/0.302/0.301 & 0.304/0.238/0.232 & fixed budget \\
Prithvi-EO-2.0 (FCN) & masked & 0,1,2 & 4 & 2048 & 200 & 181/119/130 & 0.285/0.295/0.293 & 0.246/0.263/0.256 & fixed budget \\
\quad U-Net decoder & masked & 0 & 4 & 2048 & 200 & 187 & 0.279 & 0.246 & fixed budget \\
DOFA (U-Net dec.) & masked & 0,1,2 & 4 & 2048 & 200 & 163/193/189 & 0.275/0.267/0.271 & 0.234/0.215/0.259 & fixed budget \\
TerraMind (U-Net dec.) & masked & 0,1,2 & 4 & 2048 & 200 & 165/141/149 & 0.304/0.296/0.298 & 0.242/0.268/0.245 & fixed budget \\
\quad LR $5\times10^{-5}$ & masked & 0 & 4 & 2048 & 200 & 165 & 0.296 & 0.229 & fixed budget \\
MiT-B2 U-Net & masked & 0,1,2 & 4 & 2048 & 200 & 101/180/139 & 0.300/0.302/0.288 & 0.298/0.253/0.251 & fixed budget \\
\quad original split & original & 0,1,2 & 4 & 2048 & 200 & 170/110/199 & 0.326/0.310/0.300 & 0.239/0.294/0.300 & fixed budget \\
\quad block-drop split & block drop & 0,1,2 & 4 & 2048 & 200 & 162/178/143 & 0.283/0.275/0.275 & 0.235/0.260/0.249 & fixed budget \\
ResNet-50 U-Net & masked & 0,1,2 & 4 & 2048 & 200 & 177/159/198 & 0.255/0.258/0.281 & 0.241/0.231/0.260 & fixed budget \\
\midrule
\multicolumn{10}{l}{\textit{Runs of the submitted manuscript (Phase E, original split)}} \\
Clay v1.0 (FCN) & original & - & 4 & 2048 & 114 (150) & 73 & 0.318 & - & early stop \\
Clay v1.5 (FCN) & original & - & 4 & 2048 & 150 (150) & 136 & 0.302 & - & max.\ epochs \\
Prithvi-EO-2.0 (FCN) & original & - & 4 & 2048 & 79 (150) & 38 & 0.253 & - & early stop \\
DOFA (U-Net dec.) & original & - & 4 & 2048 & 150 (150) & 125 & 0.308 & - & max.\ epochs \\
TerraMind (U-Net dec.) & original & - & 4 & 2048 & 98 (150) & 57 & 0.298 & - & early stop \\
MiT-B2 U-Net & original & - & 8 & 4096 & 150 (150) & 114 & 0.302 & - & max.\ epochs \\
\bottomrule
\end{tabular}
\end{threeparttable}
\end{adjustwidth}
\noindent{\footnotesize{Best epoch counted from 0 (as in the checkpoint names); Epochs = epochs run (maximum in brackets for the manuscript runs). Validation patches are a new random draw every epoch, so best and last validation IoU partly differ by draw. Masked split: the four 2019/20 test patches plus a 128\,m buffer removed from training and validation. Original: split of the submitted manuscript. Block drop: every grid block touching a 2019/20 test patch removed.}}

\end{table}

\begin{table}[H]

\caption{\textls[-15]{Learning rates (LRs) of all finished training runs per model logged in Weights \& Biases (preliminary runs 2025/26, Phases A to G, revision; Phase E = the runs of the submitted manuscript, Table~\ref{tab:r1-b8_training_facts}). Decoders were fixed per model; the LR was chosen from the runs before Phase E, which differ in input bands, schedule, and number of epochs, so their validation scores are not directly comparable.}\label{tab:r1-b8_lr_table}}
\scriptsize
\begin{adjustwidth}{-\extralength}{0cm}

\begin{tabularx}{\fulllength}{llrm{8.2cm}cc}
\toprule
\textbf{Model} & \textbf{Decoder} & \textbf{Runs} & \textbf{LRs Explored Before Phase E (Runs)} & \textbf{LR Phase E} & \textbf{LR Revision} \\
\midrule
Clay v1.0 & FCN/U-Net & 29 & $2.5\times10^{-5}$ (17), $5\times10^{-5}$ (1), $7.5\times10^{-5}$ (1), $10^{-4}$ (2), $10^{-3}$ (1) & $2.5\times10^{-5}$ & $2.5\times10^{-5}$ \\
Clay v1.5 & FCN & 15 & $2.5\times10^{-5}$ (3), $5\times10^{-5}$ (1), $10^{-4}$ (5) & $2.5\times10^{-5}$ & $2.5\times10^{-5}$ \\
Prithvi-EO-2.0 & FCN/U-Net & 10 & $5\times10^{-5}$ (3), $10^{-4}$ (1) & $5\times10^{-5}$ & $5\times10^{-5}$ \\
DOFA & U-Net & 11 & $5\times10^{-5}$ (2), $10^{-4}$ (2) & $10^{-4}$ & $10^{-4}$ \\
TerraMind & U-Net & 11 & $10^{-5}$ (2), $6\times10^{-5}$ (1), $10^{-4}$ (1) & $6\times10^{-5}$ & $6\times10^{-5}$ \\
MiT-B2 U-Net & U-Net & 7 & - & $10^{-4}$ & $10^{-4}$ \\
ResNet-50 U-Net & U-Net & 3 & - & - & $10^{-4}$ \\
\bottomrule
\end{tabularx}

\end{adjustwidth}
\noindent{\footnotesize{U-Net decoder for Clay v1.0 and Prithvi-EO-2.0 only in the revision decoder check; TerraMind revision includes one check run at $5\times10^{-5}$. No runs before Phase E are logged for the MiT-B2 U-Net; the ResNet-50 U-Net was added in the revision. Runs: all finished runs incl. Phases E to G and the revision.}}

\end{table}
\vspace{-6pt}
\begin{table}[H]

\caption{Best validation checkpoint versus last epoch: macro IoU at 0.2\,m on the 2019/20 test patches, mean $\pm$ standard deviation over training seeds; masked split unless stated. Seg.: single-date segmentation (23 classes); CD: 2013$\rightarrow$2019/20 change detection of the model output (before post-processing). $\Delta$: best minus last, mean over the seeds that have both. The best checkpoint is used in all main tables. Validation draws a new random set of 256 patches every epoch, so best-epoch selection is noisy; best minus last in segmentation ranges from $-$0.007 to +0.009 per model (CD multi-class $-$0.005 to +0.015, CD binary $-$0.015 to +0.010).\label{tab:r1-b9_best_vs_last}}
\scriptsize
\begin{adjustwidth}{-\extralength}{0cm}
\setlength{\tabcolsep}{5.2pt}
\centering
\begin{threeparttable}
\begin{tabular}{lccrccrccr}
\toprule
& \multicolumn{3}{c}{\textbf{Seg.\ 2019/20}} & \multicolumn{3}{c}{\textbf{CD Multi-Class}} & \multicolumn{3}{c}{\textbf{CD Binary}} \\
\cmidrule{2-10}
\textbf{Model} & \textbf{Best} & \textbf{Last} & \boldmath{$\Delta$} & \textbf{Best} & \textbf{Last}& \boldmath{$\Delta$}  & \textbf{Best} & \textbf{Last} & \boldmath{$\Delta$} \\
\midrule
Clay v1.0 & 0.249 $\pm$ 0.003 & 0.251 $\pm$ 0.001 & $-$0.001 & 0.233 $\pm$ 0.003 & 0.236 $\pm$ 0.001 & $-$0.003 & 0.425 $\pm$ 0.007 & 0.427 $\pm$ 0.004 & $-$0.002 \\
Clay v1.5 & 0.210 $\pm$ 0.006 & 0.207 $\pm$ 0.005 & +0.003 & 0.209 $\pm$ 0.005 & 0.208 $\pm$ 0.005 & +0.001 & 0.386 $\pm$ 0.011 & 0.385 $\pm$ 0.010 & +0.001 \\
Prithvi-EO-2.0 & 0.216 $\pm$ 0.013 & 0.215 $\pm$ 0.004  $^{a}$ & $-$0.006 & 0.204 $\pm$ 0.009 & 0.202 $\pm$ 0.011  $^{a}$ & +0.001 & 0.394 $\pm$ 0.012 & 0.403 $\pm$ 0.003  $^{a}$ & $-$0.015 \\
DOFA & 0.201 $\pm$ 0.004 & 0.202 $\pm$ 0.005 & $-$0.001 & 0.210 $\pm$ 0.013 & 0.209 $\pm$ 0.011 & +0.001 & 0.395 $\pm$ 0.009 & 0.396 $\pm$ 0.007 & $-$0.001 \\
TerraMind & 0.208 $\pm$ 0.003 & 0.208 $\pm$ 0.001 & 0.000 & 0.204 $\pm$ 0.005 & 0.201 $\pm$ 0.004 & +0.002 & 0.388 $\pm$ 0.007 & 0.394 $\pm$ 0.004 & $-$0.005 \\
MiT-B2 U-Net & 0.226 $\pm$ 0.015 & 0.232 $\pm$ 0.003 & $-$0.005 & 0.209 $\pm$ 0.018 & 0.210 $\pm$ 0.008 & 0.000 & 0.389 $\pm$ 0.020 & 0.391 $\pm$ 0.004 & $-$0.002 \\
ResNet-50 U-Net & 0.157 $\pm$ 0.014 & 0.156 $\pm$ 0.014 & +0.001 & 0.164 $\pm$ 0.014 & 0.164 $\pm$ 0.009 & 0.000 & 0.344 $\pm$ 0.017 & 0.339 $\pm$ 0.012 & +0.005 \\
\midrule
\multicolumn{10}{l}{\textit{Checks and controls}} \\
Clay v1.0, U-Net decoder & 0.224 & 0.215 & +0.009 & 0.227 & 0.212 & +0.015 & 0.408 & 0.398 & +0.010 \\
Prithvi-EO-2.0, U-Net decoder & 0.202 & 0.203 & $-$0.002 & 0.221 & 0.218 & +0.002 & 0.396 & 0.398 & $-$0.002 \\
TerraMind, LR $5\times10^{-5}$ & 0.219 & 0.213 & +0.005 & 0.214 & 0.212 & +0.001 & 0.400 & 0.401 & $-$0.002 \\
MiT-B2 U-Net, original split & 0.228 $\pm$ 0.002 & 0.234 $\pm$ 0.007 $^{a}$ & $-$0.007 & 0.183 $\pm$ 0.012 & 0.190 $\pm$ 0.009 $^{a}$ & $-$0.005 & 0.385 $\pm$ 0.010 & 0.390 $\pm$ 0.003  $^{a}$ & $-$0.007 \\
MiT-B2 U-Net, block-drop split & 0.215 $\pm$ 0.006 & 0.210 $\pm$ 0.004  $^{a}$ & +0.002 & 0.203 $\pm$ 0.012 & 0.203 $\pm$ 0.001 $^{a}$ & $-$0.005 & 0.378 $\pm$ 0.006 & 0.374 $\pm$ 0.002  $^{a}$ & +0.001 \\
\bottomrule
\end{tabular}
\end{threeparttable}
\end{adjustwidth}

\noindent{\footnotesize{\textsuperscript{a} Two seeds: seed 0 of this run has no last-epoch raster.}}

\end{table}
\vspace{-6pt}
\begin{table}[H]

\caption{Model size and computational cost (seed-0 revision runs, best checkpoint). Inference: one NVIDIA RTX 4000 Ada (20\,GB), fp32, batch size 4, model forward pass, and argmax only on 5$\times$256$\times$256 patches, median of 7$\times$20 batches after 10 warm-up batches. Training: 200 epochs of 2048 patches, batch size 4, 16-bit mixed precision. All parameters are trained (no frozen encoders).\label{tab:r1-b10_efficiency}}
\scriptsize
\begin{adjustwidth}{-\extralength}{0cm}
\setlength{\tabcolsep}{2.6pt}
\centering
\begin{threeparttable}
\begin{tabular}{lrrr c rrrr c lrr >{\raggedright\arraybackslash}p{3.2cm}}
\toprule
& \multicolumn{3}{c}{\textbf{Parameters (M)}} & & \multicolumn{4}{c}{\textbf{Inference (RTX 4000 Ada)}} & & \multicolumn{3}{c}{\textbf{Training (200 Epochs)}} & \\
\midrule
\textbf{Model }& \textbf{Total} & \textbf{Enc. }& \textbf{Dec.} & & \textbf{GFLOPs \textsuperscript{a}} &\textbf{ Patches/s} & \textbf{Mem.\ (GiB) \textsuperscript{b}} & \textbf{s/km$^2$ \textsuperscript{c}} & & \textbf{GPU \textsuperscript{d}} & \textbf{Time (h)} & \textbf{Mem.\ (GB) \textsuperscript{e}} & \textbf{Pretraining} \\
\midrule
MiT-B2 U-Net & 27.5 & 24.2 & 3.3 &  & 15 & 341 & 0.2 & 0.7 &  & RTX 4000 & 3.2 & 1.8 & MiT-B2 (SegFormer), ImageNet \\
ResNet-50 U-Net & 32.5 & 23.5 & 9.0 &  & 22 & 541 & 0.4 & 0.4 &  & RTX 4000 & 3.2 & 1.8 & ResNet-50, ImageNet \\
Clay v1.0 (FCN) & 96.0 & 92.1 & 3.9 &  & 674 & 17.9 & 4.5 & 12.6 &  & H200 & 14.0 & 13.8 & ViT-B, satellite + sub-metre aerial (NAIP, LINZ) \\
Clay v1.5 (FCN) & 315.6 & 311.4 & 4.2 &  & 1184 & 12.6 & 5.7 & 17.8 &  & H200 & 16.5 & 20.6 & ViT-L, satellite + sub-metre aerial (NAIP, LINZ) \\
Prithvi-EO-2.0 300M (FCN) & 308.4 & 303.6 & 4.8 &  & 355 & 50.3 & 2.2 & 4.5 &  & H200 & 5.9 & 10.3 & ViT-L, 4.2M HLS samples (30\,m) \\
DOFA base (U-Net dec.) & 126.9 & 111.4 & 15.5 &  & 65 & 276 & 0.9 & 0.8 &  & RTX 4000 & 5.2 & 3.4 & ViT-B, five sensors incl. NAIP ($\sim$1\,m) \\
TerraMind base (U-Net dec.) & 101.5 & 85.9 & 15.5 &  & 65 & 307 & 0.5 & 0.7 &  & RTX 4000 & 5.1 & 3.2 & ViT-B, generative multimodal (incl. S2, 30\,m DEM) \\\bottomrule
\end{tabular} %
\end{threeparttable}
\end{adjustwidth}
\end{table}

\begin{table}[H]\ContinuedFloat
\caption{\textit{Cont.}}
\scriptsize
\begin{adjustwidth}{-\extralength}{0cm}
\setlength{\tabcolsep}{2.12pt}
\centering
\begin{threeparttable}
\begin{tabular}{lrrr c rrrr c lrr >{\raggedright\arraybackslash}p{3.2cm}}
\toprule
& \multicolumn{3}{c}{\textbf{Parameters (M)}} & & \multicolumn{4}{c}{\textbf{Inference (RTX 4000 Ada)}} & & \multicolumn{3}{c}{\textbf{Training (200 Epochs)}} & \\
\midrule
\textbf{Model }& \textbf{Total} & \textbf{Enc. }& \textbf{Dec.} & & \textbf{GFLOPs \textsuperscript{a}} &\textbf{ Patches/s} & \textbf{Mem.\ (GiB) \textsuperscript{b}} & \textbf{s/km$^2$ \textsuperscript{c}} & & \textbf{GPU \textsuperscript{d}} & \textbf{Time (h)} & \textbf{Mem.\ (GB) \textsuperscript{e}} & \textbf{Pretraining} \\
\midrule

\multicolumn{14}{l}{\textit{Decoder check (one seed)}} \\
Clay v1.0 (U-Net dec.) & 107.6 & 92.1 & 15.5 &  & 287 & 61.8 & 1.4 & 3.6 &  & RTX 4000 & 11.2 & 6.0 & as above \\
Prithvi-EO-2.0 300M (U-Net dec.) & 323.9 & 303.6 & 20.3 &  & 185 & 113 & 2.6 & 2.0 &  & H200 & 5.0 & 8.1 & as above \\
\bottomrule
\end{tabular}

\end{threeparttable}
\end{adjustwidth}

\noindent{\footnotesize{\textsuperscript{a} Per forward pass of one patch, counted with \texttt{torch.utils.flop\_counter} (one multiply-add = 2 FLOPs, attention included). \textsuperscript{b} Peak \texttt{torch.cuda.max\_memory\_allocated} incl.\ weights, under \texttt{torch.inference\_mode}. \textsuperscript{c} Model compute per km$^2$ at 0.5\,m with the inference tiling (256\,px patches, 64\,px overlap per side, 225 patches per km$^2$); excludes reading and writing data, which dominated the actual pipeline.  \textsuperscript{d} Wall time is comparable only within one machine: the RTX 4000 Ada was shared by two runs, the H200 by three revision runs and other users' jobs. \textsuperscript{e} Peak training memory from 2-epoch test runs at batch size 4; Prithvi-EO-2.0 (FCN) measured on the RTX 4000 Ada, Clay v1.0 (FCN) and Clay v1.5 on the H200.}}

\end{table}

\begin{adjustwidth}{-\extralength}{0cm}

\reftitle{References}

\PublishersNote{}
\end{adjustwidth}
\end{document}